\documentclass{article}

\usepackage{microtype}
\usepackage{graphicx}
\usepackage{subfig}
\usepackage{booktabs} 

\usepackage{hyperref}

\usepackage[accepted]{icml2024arxiv}

\usepackage{amsmath}
\usepackage{amssymb}
\usepackage{mathtools}
\usepackage{amsthm}

\usepackage[capitalize,noabbrev]{cleveref}

\theoremstyle{plain}
\newtheorem{theorem}{Theorem}[section]

\newtheorem{lemma}[theorem]{Lemma}
\newtheorem{corollary}[theorem]{Corollary}
\theoremstyle{definition}

\theoremstyle{remark}

\usepackage[textsize=tiny]{todonotes}

\usepackage{graphicx}

\usepackage{booktabs}
\usepackage{multirow}

\usepackage{adjustbox}

\usepackage{framed}

\usepackage{dsfont}
\usepackage{amsfonts}
\usepackage{amsmath}
\usepackage{amsthm}
\usepackage{mathtools}
\DeclareMathOperator*{\argmax}{arg\,max}
\DeclareMathOperator*{\argmin}{arg\,min}
\usepackage{xurl}

\usepackage[english]{babel}

\newcommand{\BOS}{\textsc{BOS}}
\newcommand{\EOS}{\textsc{EOS}}
\newcommand{\vx}{\mathbf{x}}
\newcommand{\vy}{\mathbf{y}}
\newcommand{\vh}{\mathbf{h}}

\usepackage{xcolor}

\newcommand{\baseline}{MBR}
\newcommand{\weighted}{MBMBR}
\newcommand{\lnw}{MBMBR\textsubscript{L}}

\newcommand{\Pmodel}{P_\mathrm{{model}}} 
\newcommand{\Pm}{P}
\newcommand{\Phat}{\hat{\Pm}}
\newcommand{\Pmb}{\Phat_{\mathrm{MB}}}

\newcommand{\CandH}{\mathcal{H}_{\mathrm{cand}}}
\newcommand{\RefH}{\mathcal{H}_{\mathrm{ref}}}

\icmltitlerunning{Model-Based Minimum Bayes Risk Decoding for Text Generation}

\begin{document}

\twocolumn[
\icmltitle{Model-Based Minimum Bayes Risk Decoding for Text Generation}



\icmlsetsymbol{equal}{*}

\begin{icmlauthorlist}
\icmlauthor{Yuu Jinnai}{ca}
\icmlauthor{Tetsuro Morimura}{ca}
\icmlauthor{Ukyo Honda}{ca}
\icmlauthor{Kaito Ariu}{ca}
\icmlauthor{Kenshi Abe}{ca}
\end{icmlauthorlist}

\icmlaffiliation{ca}{CyberAgent, Tokyo, Japan}

\icmlcorrespondingauthor{Yuu Jinnai}{jinnai\_yu@cyberagent.co.jp}

\icmlkeywords{Text Generation, Machine Translation, Minimum Bayes Risk Decoding}

\vskip 0.3in
]



\printAffiliationsAndNotice{}  

\begin{abstract}

Minimum Bayes Risk (MBR) decoding has been shown to be a powerful alternative to beam search decoding in a variety of text generation tasks.
MBR decoding selects a hypothesis from a pool of hypotheses that has the least expected risk under a probability model according to a given utility function.
Since it is impractical to compute the expected risk exactly over all possible hypotheses, two approximations are commonly used in MBR. 
First, it integrates over a sampled set of hypotheses rather than over all possible hypotheses. Second, it estimates the probability of each hypothesis using a Monte Carlo estimator.
While the first approximation is necessary to make it computationally feasible, the second is not essential since we typically have access to the model probability at inference time. 
We propose model-based MBR (MBMBR), a variant of MBR that uses the model probability itself as the estimate of the probability distribution instead of the Monte Carlo estimate.
We show analytically and empirically that the model-based estimate is more promising than the Monte Carlo estimate in text generation tasks.
Our experiments show that MBMBR outperforms MBR in several text generation tasks, both with encoder-decoder models and with language models.
Our code is available at \url{https://github.com/CyberAgentAILab/model-based-mbr}.
\end{abstract}

%
%


\section{Introduction}

One of the key components of text generation is the decoding strategy, which is the decision rule used to generate sentences from the model.
Beam search is the popular decoding strategy that has been widely used in many directed text generation tasks, including machine translation \cite{wu2016googles,ott-etal-2019-fairseq,wolf-etal-2020-transformers}, text summarization \cite{rush-etal-2015-neural,narayan-etal-2018-dont}, and image captioning \cite{anderson-etal-2017-guided}.
However, beam search is known to have several degeneration problems. For example, \citet{welleck-etal-2020-consistency} report that beam search can yield infinite-length outputs that the model assigns zero probability to.

\textbf{Minimum Bayes Risk (MBR) decoding} has recently gained attention as a decoding strategy with the potential to overcome the problems of beam search \cite{goodman-1996-parsing,kumar-byrne-2004-minimum,eikema-aziz-2020-map,eikema-aziz-2022-sampling,freitag-etal-2022-high,bertschs2023}.
MBR decoding consists of the following steps. First, it samples multiple sequences from the model. Then, it compares each sequence to the others according to a utility function. Finally, it selects the sequence that maximizes the expected utility over the estimated probability distribution over the sequences.

Previous work on MBR decoding uses a Monte Carlo estimate to approximate the probability distribution since it is an unbiased estimate of the true model distribution.
However, because the number of possible hypotheses is enormous compared to the number of samples, the estimation error of the Monte Carlo estimate is huge. This can lead to a huge error in the expected utility estimate.

We propose \textbf{model-based MBR (MBMBR)}, a variant of MBR that uses a model-based estimate of the MBR objective instead of a Monte Carlo estimate. The model-based estimate uses the probability model itself, but with its domain limited to the set of observed hypotheses. As such, the estimate is computationally feasible and as accurate as the model for the observed sequences. MBMBR is easy to implement and requires only the model probability of the sampled sequences, which can be obtained concurrently with the sampling procedure.

We first evaluate the model-based estimate and show analytically that the Kullback-Leiber (KL) divergence from the true model probability is guaranteed to be lower than that of the Monte Carlo estimate.
We also empirically evaluate it on various text generation tasks, indicating that model-based estimate is better for the use of MBR.

We then apply MBMBR to several text generation tasks: machine translation, text summarization, image captioning, and data-to-text. We evaluate it in two settings: using a domain-specific conditional generation model, and using a language model (LLM) with prompting.
MBMBR outperforms MBR in all settings except when the model generates low-quality sequences.
The experimental results show that MBMBR is an effective decoding strategy that can replace MBR to improve text generation in a wide range of settings.

\section{Background}
\label{sec:text}
Text generation is the task of generating an output sequence $\vy$ given an input sequence $\vx$.
Probabilistic text generators define a probability distribution $\Pmodel (\vy | \vx)$ over an output space of hypotheses $\mathcal{Y}$. 
The set of complete hypotheses $\mathcal{Y}$ is:
\begin{equation}
    \mathcal{Y} := \{\BOS \circ \mathbf{v} \circ \EOS | \mathbf{v} \in \mathcal{V}^*\},
\end{equation}
where $\circ$ is a string concatenation and $\mathcal{V}^*$ is the Kleene closure of a set of vocabulary $\mathcal{V}$. 
The goal of decoding is to find the highest-scoring hypothesis for a given input. 

One of the most common decision rules is maximum-a-posteriori (MAP) decoding. MAP decoding finds the most probable translation under the model:
\begin{equation}
    \vh^{\mathrm{MAP}} = \argmax_{\vh \in \mathcal{Y}} \Pmodel(\vh | \vx).
\end{equation}
In this paper, we denote $\Pmodel(\vh | \vx)$ by $\Pmodel(\vh)$ for brevity.
Although it seems intuitive to solve this MAP objective, previous work has pointed out two critical problems with this strategy. First, because the size of the hypothesis set $|\mathcal{Y}|$ is extremely large, it is intractable to solve exactly. Second, the MAP objective often leads to low quality output \cite{stahlberg-byrne-2019-nmt,Holtzman2020The,meister-etal-2020-beam}. Indeed, \citet{stahlberg-byrne-2019-nmt} show that $\vh^{\mathrm{MAP}}$ is often found to be the empty sequence in their experimental setting.

As such, beam search is commonly used as a heuristic algorithm to solve a decoding problem \cite{graves2012sequence,Sutskever2014}. 
Beam search considers a fixed number $k$ of options at each step. 
It is known to produce higher quality sequences than MAP decoding in a wide range of tasks.
However, beam search is known to have degeneration problems such as repetitions and infinite length outputs \cite{pmlr-v97-cohen19a,Holtzman2020The}.

\subsection{Minimum Bayes Risk Decoding}

\begin{table*}
    \centering
    \begin{tabular}{llcl}
        \toprule
        \multicolumn{2}{c}{Method} & Estimator & \multicolumn{1}{c}{Objective} \\
        \cmidrule(lr){1-2}
        \cmidrule(lr){3-3}
        \cmidrule(lr){4-4}
        Target Objective & $\vh^{\mathrm{model}}$ (Eq.~\ref{eq:mbr}) & $\Pm$ & $\sum_{\vy \in \mathcal{Y}} u(\vh, \vy) \Pm(\vy) $ \\
        Monte Carlo (MBR) & $\vh^{\mathrm{MC}}$ (Eq.~\ref{eq:empirical}) & $\Phat$ & $\sum_{\vy \in \textcolor{red}{\RefH}} u(\vh, \vy) \textcolor{red}{\hat{\Pm}}(\vy) $ \\
        Model-Based (MBMBR) & $\vh^{\mathrm{MB}}$ (Eq.~\ref{eq:mbmbr}) & $\Pmb$ & $\sum_{\vy \in \textcolor{red}{\mathcal{R}}} u(\vh, \vy) \Pm(\vy)$ \\
        \bottomrule
    \end{tabular}
    \caption{Comparison of the surrogate objective functions of MBMBR and MBR. MBR uses Monte Carlo estimate to approximate the target MBR objective whereas MBMBR uses the model $\Pm$ as is.}
    \label{tab:mbr-obj}
\end{table*}

Unlike MAP decoding, which searches for the most probable output, MBR decoding searches for the output that maximizes expected utility, which is equivalent to minimizing risk \cite{goel2000minimum,kumar-byrne-2002-minimum,kumar-byrne-2004-minimum}.
The procedure consists of two components: a text generation model $\Pmodel(\vy)$ and a utility metric $u(\vh, \vy)$.\footnote{We denote $\Pmodel(\vy | \vx)$ by $\Pmodel(\vy)$ for simplicity.}
The utility metric $u(\vh, \vy)$ estimates the quality of a candidate output $\vh$ given a reference output $\vy$.
Given a set of candidate hypotheses $\CandH \subseteq \mathcal{Y}$, MBR decoding selects the best hypothesis according to its expected utility over the distribution of human references:
\begin{equation}
    \vh^{\mathrm{human}} = \argmax_{\vh \in \CandH}  \mathop{\mathbb{E}}_{\vy \sim {P_\mathrm{human}}} [u(\vh, \vy)].
\end{equation}
Since $P_\mathrm{human}$ is unknown, MBR instead uses the model probability $\Pmodel$ to approximate $P_\mathrm{human}$.
\begin{align}
    \vh^{\mathrm{model}} &= \argmax_{\vh \in \CandH} \mathop{\mathbb{E}}_{\vy \sim \Pmodel} [u(\vh, \vy)] \nonumber \\
     & =  \argmax_{\vh \in \CandH} \sum_{\vy \in \mathcal{Y}} u(\vh, \vy) \cdot \Pmodel(\vy).  
\label{eq:mbr}
\end{align}
For the rest of the paper, we will denote $\Pmodel$ as $\Pm$ for simplicity, unless otherwise noted.
Since integration over $\mathcal{Y}$ is computationally intractable, Eq.~\eqref{eq:mbr} is approximated by a \textbf{Monte Carlo estimate} \cite{eikema-aziz-2022-sampling,farinhas2023empirical} using a collection of reference hypotheses $\RefH$ sampled from the model $\Pm$:\footnote{$\RefH$ is a collection where it may have multiple instances of the same element.}
\begin{align}
    \vh^{\mathrm{MC}} &= \argmax_{\vh \in \CandH} \frac{1}{|\RefH|} \sum_{\vy \in \RefH} u(\vh, \vy).
\label{eq:empirical-p}
\end{align}
Eq.~\eqref{eq:empirical-p} is derived by replacing the true model probability $\Pm$ in Eq.~\eqref{eq:mbr} with the \textbf{empirical distribution} $\Phat$, which is the number of occurrences of $\vy$ in $\RefH$ divided by the number of samples $|\RefH|$:
\begin{equation}
    \hat{\Pm}(\vy) = \frac{\sum_{\vy' \in \RefH} \mathbb{I}(\vy = \vy')}{|\RefH|}.
\end{equation}
Using $\Phat$, we can rewrite Eq.~\eqref{eq:empirical-p} as follows:
\begin{align}
    \vh^{\mathrm{MC}} &= \argmax_{\vh \in \CandH} \mathop{\mathbb{E}}_{\vy \sim \hat{\Pm}} [u(\vh, \vy)] \nonumber \\
    &= \argmax_{\vh \in \CandH} \sum_{\vy \in \mathcal{Y}} u(\vh, \vy) \cdot \Phat(\vy) \nonumber \\
    &= \argmax_{\vh \in \CandH} \sum_{\vy \in \RefH} u(\vh, \vy) \cdot \hat{\Pm}(\vy). 
\label{eq:empirical}
\end{align}
Note that $\Phat(\vy)$ is zero for $\vy \in \mathcal{Y} \setminus \RefH$.
Standard practice is to use the same set of hypotheses for the candidate set $\CandH$ and the reference pool $\RefH$ ($\mathcal{H} \coloneqq \CandH = \RefH$).

The choice of sampling algorithm to collect $\mathcal{H}$ has been studied extensively, as it has been shown to be critical to the performance of MBR \cite{ohashi2024true}. Recent work has shown that MBR works best with a probabilistic sampling algorithm \cite{eikema-aziz-2020-map,fernandes-etal-2022-quality,freitag2023epsilon}.

\section{Model-Based Minimum Bayes Risk (MBMBR) Decoding}

The estimation error of the Monte Carlo sum comes from two approximations.
First, the domain of the reference hypotheses is restricted to the collection of sampled sentences $\RefH$ instead of $\mathcal{Y}$. This approximation is necessary because enumerating all hypotheses in $\mathcal{Y}$ is infeasible. Second, the probability of each sentence in $\RefH$ is estimated by the Monte Carlo estimate instead of the true model probability $\Pm$.
While the first approximation is necessary, the second approximation is unnecessary if the model probability is accessible, which is the case for most decoding scenarios.
To this end, we propose a \textbf{model-based MBR} (MBMBR) decoding, a variant of MBR that uses a \textbf{model-based estimate} instead of a Monte Carlo estimate:
\begin{equation}
    \vh^{\mathrm{MB}} = \argmax_{\vh \in \CandH} \sum_{\vy \in \mathcal{R}} u(\vh, \vy) \cdot \Pm(\vy),
\label{eq:mbmbr}
\end{equation}
where $\mathcal{R}$ is a set of references where the duplicates are removed from the $\RefH$.
The model-based estimate simply replaces $\hat{\Pm}$ with $\Pm$.
While vanilla MBR estimates the probability distribution using only the samples, MBMBR uses both the samples and their model probability to estimate. 
Intuitively, MBMBR can fully exploit the inherent properties of the given model. This allows a more accurate probability density to be computed, resulting in more accurate estimates.

\begin{table*}
    \centering
    \adjustbox{max width=\textwidth}{
    \begin{tabular}{ccccc}
        \toprule
        \multicolumn{2}{c}{Sampled Texts} & Target & Monte Carlo Estimate & Model-Based Estimate \\ 
        \cmidrule(l){1-2}
        Text & \#Occurrences & $\Pm$ & $\hat{\Pm}$ & $\hat{\Pm}_{\mathrm{MB}}$ \\
        \midrule
        {\it But telling the truth is not a crime.} & 2 & 0.3 & 0.4 & 0.6 \\
        {\it However, telling the truth is not a crime.} & 2 & 0.1 & 0.4 & 0.2 \\
        {\it But to tell the truth is not a crime.} & 1 & 0.1 & 0.2 & 0.2 \\
        (All others) & 0 & 0.5 & 0 & 0 \\\midrule
        \multicolumn{2}{c}{$D_{\mathrm{KL}}(\cdot || \Pm)$} & 0 & 0.808 & \textbf{0.693} \\
        \bottomrule
    \end{tabular}
    }
    \caption{Illustrative example of the model-based estimate. While the Monte Carlo estimate $\hat{\Pm}$ uniformly assigns the same weight to the sampled text without considering the $\Pm$, the model-based estimate takes advantage of the fact that the $\Pm$ is given in the text generation task and uses $\Pm$ to adjust the distribution accordingly.}
    \label{tab:example}
\end{table*}

The idea of using model probability for MBR is not new per se. The model probability $\Pm$ is used when $\RefH$ is collected by a beam search instead of probabilistic sampling algorithms \cite{goel2000minimum,kumar-byrne-2002-minimum,kumar-byrne-2004-minimum}. 
The novelty of the idea is to show that we can make use of the model probability with a probabilistic sampling algorithm to further improve the MBR decoding. In the following sections, we show that (1) the divergence of the estimated distribution is guaranteed to be smaller (or equal in extremely unlikely cases) than using the Monte Carlo estimate (Section \ref{sec:analysis}), (2) the divergence of the estimated distribution is reduced so that about half as many samples are needed to achieve the same divergence as with Monte Carlo estimate (Figures \ref{fig:aed-vis} and \ref{fig:apx-kl}), and (3) using the model probability, MBMBR outperforms MBR in a wide range of text generation tasks (Section \ref{sec:experiments}).

\paragraph{Mitigating the Length Bias.}
The problem of using $\Pm$ directly for \weighted{} is that it imposes a bias on the hypothesis selection if the model is biased. Length bias is one of the well-known problems that text generation model tends to generate outputs that are too short \cite{koehn-knowles-2017-six,murray-chiang-2018-correcting,stahlberg-byrne-2019-nmt}.
This is problematic since $P_\mathrm{human}$ is unlikely to have such a bias as strong as the model probability $\Pm$. 
Although previous work shows that vanilla MBR decoding generates outputs relatively close in length to the references \cite{bertschs2023}, MBMBR is directly subject to the length bias since it uses the probability itself (See Section \ref{sec:sumsum}).

To remedy this problem, we use length normalization with a $\Pm$ normalized by the length of the sequence \cite{murray-chiang-2018-correcting,bertschs2023}.
Suppose the model probability is biased to generate shorter sequences: $\Pm(\vy) \sim  e^{-\ell(\vy)} P_\mathrm{human}(\vy)$, where $\ell(\vy)$ is the length of $\vy$, we compensate for this bias as follows:
\begin{equation}
    \vh^{\mathrm{MB_{L}}} = \argmax_{\vh \in \CandH} \sum_{\vy \in \mathcal{R}} u(\vh, \vy) \cdot e^{\ell(\vy)} \Pm(\vy).
\label{eq:lnadj}
\end{equation}
In this way, we expect to mitigate the assumed length bias as:
\begin{equation}
    e^{l(y)} P(y) \sim e^{l(y)} e^{-l(y)} P_\mathrm{human}(y) = P_\mathrm{human}(y).
\end{equation}
We denote a variant of MBMBR computing $\vh^{\mathrm{MB_{L}}}$ as length-normalized MBMBR (\lnw{}). 


\section{Properties of the Model-Based Estimate}
\label{sec:analysis}

In this section, we first introduce the model-based estimator $\Pmb$ for the sake of analysis and give an example to illustrate its difference from the Monte Carlo estimator (Table \ref{tab:example}). We then evaluate its divergence from the true model probability analytically and empirically.

\subsection{Probability Distribution of Model-Based Estimate}

MBMBR can be understood as a standard MBR, but using the following probability distribution instead of the empirical distribution:
\begin{equation}
    \Pmb(\vy) = \begin{cases}
        \alpha \Pm(\vy) & \text{if $\vy \in \mathcal{R}$} \\
        0      & \text{otherwise},
    \end{cases}
\label{eq:mbp}
\end{equation}
where $\alpha = \frac{1}{\sum_{\vy' \in \mathcal{R}} \Pm(\vy')}$ is a normalization factor so that it sums to 1: $\sum_{\vy \in \mathcal{Y}} \Pmb(\vy) = 1$. 
Note that $\Pmb$ is introduced solely for analysis and computing it is not necessary to run MBMBR. One can follow Eq.~\eqref{eq:mbmbr} to implement MBMBR.
$\Pmb$ is a probability distribution with support restricted to $\mathcal{R}$ and its probability is proportional to $\Pm$.
We denote $\hat{\Pm}_{\mathrm{MB}}(\vy)$ as a \textbf{model-based distribution}.

Using $\Pmb$, we can rewrite the form of the MBMBR as follows:
\begin{align}
    \vh^{\mathrm{MB}} &= \argmax_{\vh \in \CandH} \sum_{\vy \in \mathcal{R}} u(\vh, \vy) \cdot \Pm(\vy) \nonumber \\
    & =  \argmax_{\vh \in \CandH} \sum_{\vy \in \mathcal{Y}} u(\vh, \vy) \cdot \Pmb(\vy).
\end{align}
Compared to $\vh^{\mathrm{model}}$ (Eq.~\ref{eq:mbr}), the difference is that the probability distribution $\Pm$ is replaced by the model-based distribution $\Pmb$.
Therefore, as $\Pmb$ gets closer to $\Pm$, we expect $\vh^{\mathrm{MB}}$ to get closer to $\vh^{\mathrm{model}}$, which is the gold standard for the MBR objective (See Section \ref{sec:kl-bleu}).

\paragraph{Example.}
Table \ref{tab:example} describes an example to illustrate the property of model-based distribution. Suppose we sample five sentences $\RefH = (\vy_0, \vy_0, \vy_1, \vy_1, \vy_2)$ and their probabilities according to the model $\Pm$ are ${\vy_0 = 0.3, \vy_1 = 0.1}$, and ${\vy_2 = 0.1}$, respectively, and thus ${\sum_{\vy \in \mathcal{Y} \setminus \RefH} \Pm(\vy) = 0.5}$. The empirical distribution takes into account of the number of occurrences to compute the Monte Carlo sum to get ${\hat{\Pm}(\vy_0)=0.4, \hat{\Pm}(\vy_1)=0.4, \hat{\Pm}(\vy_2)=0.2}$. The model-based distribution ignores the number of occurrences and uses $\Pm$ to weight the sample, thus ${\Pmb(\vy_0)=0.6, \Pmb(\vy_1)=0.2, \Pmb(\vy_2)=0.2}$. 
The Kullback–Leibler (KL) divergence \cite{cover1999elements} of the model-based distribution from the model probability is ${D_{\mathrm{KL}}(\Pmb || \Pm) = 0.693}$ while that of the empirical distribution is ${D_{\mathrm{KL}}(\Phat || \Pm) = 0.808}$. 
This shows that the model-based distribution is a better estimate of the true model probability than the Monte Carlo estimate.


\subsection{Divergence of the Model-Based Estimate to the True Distribution}
\label{sec:analyze-div}

\paragraph{Analytical Results.} The model-based estimate is guaranteed to be better than the Monte Carlo estimate with respect to the KL divergence. 
In fact, it is optimal over a collection of probability distributions with support restricted to $\RefH$. 
Let $\Delta(\mathcal{Y})$ be a collection of probability distributions over $\mathcal{Y}$, and $\Delta(\mathcal{Y}; \RefH)$ be a subset of $\Delta(\mathcal{Y})$, a collection of probability distributions over $\mathcal{Y}$ with their support restricted to $\RefH$:
\begin{align}
    \Delta(\mathcal{Y}; &\RefH) = \nonumber \\
    &\left\{p\in \Delta(\mathcal{Y}) ~|~ \forall \vy \in \mathcal{Y}\setminus \RefH,~ p(\vy) = 0\right\}.
\end{align}
\begin{theorem}
    The model-based distribution minimizes the Kullbuck-Leiber divergence over a collection of probability distributions with their support restricted to $\RefH$.
    \begin{equation}
        \Pmb = \argmin_{p\in \Delta(\mathcal{Y}; \RefH)} \left\{D_{\mathrm{KL}}(p || \Pm)\right\}.        
    \end{equation}
\label{th:kld}
\end{theorem}
The proof is in Appendix \ref{sec:proof}. 
The model-based estimate is thus the information projection \cite{cover1999elements} of $\Pm$ onto $\Delta(\mathcal{Y}; \RefH)$.

\begin{corollary}
     The KL divergence of the model-based estimate from the true model probability is less than or equal to that of the Monte Carlo estimate:
     \begin{equation}
        D_{\mathrm{KL}}(\hat{\Pm}_{\mathrm{MB}} || \Pm) \leq D_{\mathrm{KL}}(\hat{\Pm} || \Pm).
     \end{equation}
\label{co:kld}
\end{corollary}
The proof is immediate from the Theorem \ref{th:kld}. 
The equality holds if and only if $p(\vy) = c P(\vy)$ for some constant $c$ and for $y \in \RefH$, which is an extremely unlikely condition where the ratio of the number of occurrences of each hypothesis $\vy$ in $\RefH$ exactly matches the ratio of $\Pm(\vy)$. 

\paragraph{Simulation Study of the Divergence of the Model-Based Distribution.}
We evaluate the accuracy of model-based distribution in a controlled experiment where the true distribution is known exactly.
We use a Zipf distribution as the true distribution in this experiment \cite{piantadosi2014zipf}:
\begin{equation}
P(y) = \frac{y^{-a}}{\zeta(a)}.
\end{equation}
We set $a=2$, and limit the domain to $y \in \{0, 1, ..., 499\}$. 
We run this experiment 100 times and compute the mean and the standard deviation of the KL divergence. The mean ($\pm$ stddev) of the KL divergence of the Monte Carlo estimate is $0.183\;(\pm 0.054)$, and that of the model-based estimate is 
$0.069\;(\pm 0.016)$.
The result shows that the model-based estimate is closer to the true probability distribution than the Monte Carlo estimate, as supported by Corollary \ref{co:kld}. 

\begin{figure}[t!]
    \centering
    \subfloat[KL Divergence and the number of samples $|\RefH|$]{
        \label{fig:aed-vis-kl}
        \includegraphics[width=0.95\columnwidth]{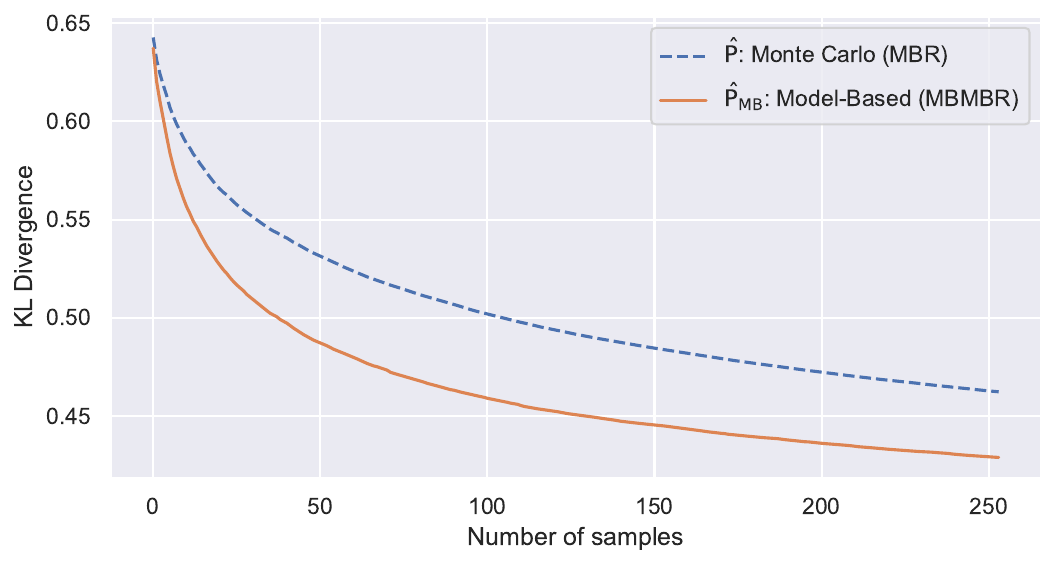}
    }\\\qquad
    \subfloat[BLEU score as a function of KL divergence]{
        \label{fig:aed-vis-bleu}
        \includegraphics[width=0.95\columnwidth]{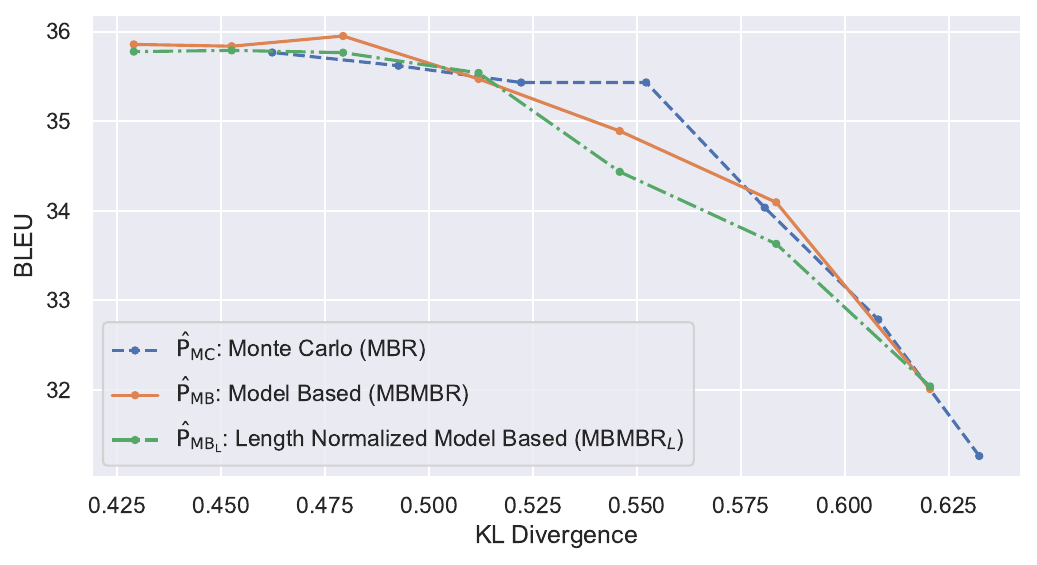}
    }
    \caption{(a) Kullback-Leibler divergence of the empirical distribution and the model-based distribution from the true model probability $\Pmodel$, averaged over the source sentences. (b) The correlation of the average KL divergence to the average BLEU score of the output. Evaluated on WMT'19 De-En.}
    \label{fig:aed-vis}
\end{figure}

\paragraph{Empirical Validation of Model-Based Estimates.}
To empirically evaluate the model-based distribution in text generation tasks, we compute the KL divergence and the Jensen-Shannon divergence on WMT'19 De-En \cite{barrault-etal-2019-findings}, IWSLT'17 Fr-En \cite{cettolo-etal-2017-overview}, XSum \cite{lewis-etal-2020-bart}, and MS COCO \cite{lin2014microsoft} datasets. We use the first 1000 inputs for each dataset. 
We sample 256 sentences for WMT'19 De-En and 64 for the other datasets using epsilon sampling ($\epsilon=0.02$). 
Figure \ref{fig:aed-vis-kl} shows the KL of the model-based and empirical distributions from $\Pm$, averaged over the source sentences of WMT'19 De-En. See Appendix \ref{sec:divergence} for the rest of the datasets and for the Jensen-Shannon divergence.
We observe that both the KL and JS divergences of the model-based distribution are significantly smaller than that of the empirical distribution in all four datasets. Approximately twice as many samples are required for the Monte Carlo estimate to achieve the same divergence as the model-based estimate.
The empirical result shows that the model-based estimate is significantly more accurate than the Monte Carlo estimate in terms of KL divergence in a wide range of tasks.

\subsection{Relationship of KL Divergence on Output Quality}
\label{sec:kl-bleu}

\paragraph{Analytical Results.}
We observe that the model-based estimate has smaller KL divergence than the Monte Carlo estimate in Section \ref{sec:analyze-div}. The question is how it affects the output of the MBR decoding.
We first show an upper bound of the estimation error of the MBR objective using KL divergence.
\begin{lemma}
    The estimation error of the MBR objective using a probability distribution $p$ as an estimate is upper bounded by the KL-divergence:
    \begin{equation}
        |\mathop{\mathbb{E}}_{\vy \sim P} [u(\vh, \vy)] - \mathop{\mathbb{E}}_{\vy \sim p} [u(\vh, \vy)]| \leq u_{\mathrm{max}} \sqrt{2 D_{\mathrm{KL}}(p || P)},
    \end{equation}
    where $u_{\mathrm{max}}$ is an upper bound of $|u|$ ($u_{\mathrm{max}} \coloneqq \max_{\vy, \vy' \in \mathcal{Y}} |u(\vy, \vy')|$).
\label{lem:pinsker}
\end{lemma}
The proof is in Appendix \ref{sec:proof2}.
The lemma shows that a lower KL divergence corresponds to a reduced upper bound on the estimation error in the MBR objective. Therefore, by minimizing the KL divergence, we anticipate generating a better hypothesis by MBR decoding.

\paragraph{Empirical Validation of the KL Divergence and the BLEU Score.}
Although Lemma \ref{lem:pinsker} shows that KL divergence upper bounds the estimation error of the MBR objective, its practical implication is unclear. To this end, we evaluate the correlation of the average BLEU scores and the average KL divergence of the model-based and Monte Carlo estimates using the WMT'19 De-En dataset. The experimental setup is the same as in Section \ref{sec:analyze-div}.
We compute the BLEU score for $|\RefH| \in \{4, 8, 16, 32, 64, 128, 256\}$.
The relationship of the KL divergence and the BLEU score is present in Figure \ref{fig:aed-vis-bleu}. The lower the KL divergence is the BLEU score tends to be higher in both algorithms. This shows that MBR using the model-based estimates achieve higher output quality (BLEU score) by reducing the KL divergence from the true model probability.


\section{Experiments}
\label{sec:experiments}

\begin{table*}[t]
    \centering
    \adjustbox{max width=\textwidth}{
    \begin{tabular}{lrrrrrrrrrrrrrr}
    \toprule
     & \multicolumn{7}{c}{WMT'19 De-En} & \multicolumn{7}{c}{WMT'19 Ru-En} \\
     & \multicolumn{7}{c}{Epsilon Sampling ($\epsilon=0.02$)}  & \multicolumn{7}{c}{Epsilon Sampling ($\epsilon=0.02$)}  \\
     \cmidrule(l){2-8} \cmidrule(l){9-15}
    $|\mathcal{H}|$ & 4 &  8 & 16 & 32 & 64 & 128 & 256 & 4 &  8 & 16 & 32 & 64 & 128 & 256 \\
    \midrule
    \small{\baseline}     & 31.26 & 32.79 & 34.04 & 35.43 & 35.43 & 35.62 &  35.77   & 28.00 & 30.03 & 31.14 & 31.95 & 31.94 &  \underline{32.68} &  32.61 \\
    \small{\weighted}   & \textbf{32.01} & \textbf{34.09} & \textbf{34.89} & \underline{35.47} & \textbf{35.95} & \underline{35.84} &  \textbf{35.86}   & \textbf{29.55} & \textbf{30.77} & \textbf{31.69} & 32.25 & \underline{32.21} &  32.58 &  \textbf{32.90}\\
    \small{\lnw} & \underline{31.96} & \underline{33.53} & \underline{34.29} & \textbf{35.77} & \underline{35.61} & \textbf{35.85} &  \underline{35.84} & \underline{28.99} & \underline{30.50} & \underline{31.40} & \textbf{32.27} & \textbf{32.27} &  \textbf{33.01} &  \underline{32.86}  \\
    \midrule
     & \multicolumn{7}{c}{Top-$k$ Sampling ($k=10$)} & \multicolumn{7}{c}{Top-$k$ Sampling ($k=10$)} \\
    \cmidrule(l){2-8} \cmidrule(l){9-15}
    $|\mathcal{H}|$ & 4 &  8 & 16 & 32 & 64 & 128 & 256 & 4 &  8 & 16 & 32 & 64 & 128 & 256 \\
    \midrule
    \small{\baseline} & 29.73 & 31.43 & 32.84 & 33.57 & 34.26 &  34.55 &  34.82 & 27.09 & 29.33 & 30.23 & 30.95 & 31.46 &  31.69 &  31.80 \\
    \small{\weighted} & \textbf{30.8}7 & \textbf{32.34} & \textbf{33.79} & \textbf{35.09} & \textbf{35.34} &  \textbf{35.62} &  \textbf{35.49} & \underline{27.96} & \textbf{29.71} & \textbf{30.75} & \textbf{31.48} & \textbf{31.81} &  \underline{31.86} &  \underline{31.93} \\
    \small{\lnw}      & \underline{30.68} & \underline{31.97} & \underline{33.35} & \underline{34.24} & \underline{34.73} &  \underline{34.78} &  \underline{35.26}  & \textbf{28.11} & \underline{29.54} & \underline{30.39} & \underline{31.26} & \underline{31.59} &  \textbf{32.11} &  \textbf{32.09} \\
    \midrule
     & \multicolumn{7}{c}{Nucleus Sampling ($p=0.9$)}  & \multicolumn{7}{c}{Nucleus Sampling ($p=0.9$)} \\
    \cmidrule(l){2-8} \cmidrule(l){9-15}
    $|\mathcal{H}|$ & 4 &  8 & 16 & 32 & 64 & 128 & 256 & 4 &  8 & 16 & 32 & 64 & 128 & 256 \\
    \midrule
    \small{\baseline} & 28.71 & 30.35 & 31.90 & 33.46 & \underline{34.13} &  34.69 &  34.80  & 25.51 & 28.46 & 29.25 & 30.15 & 30.79 &  31.43 &  31.35  \\
    \small{\weighted} & \textbf{29.63} & \textbf{31.49} & \textbf{32.43} & \textbf{33.88} & \textbf{34.14} &  \textbf{34.89} &  \textbf{35.49} & \underline{26.65} & \underline{28.47} & \underline{29.40} & \textbf{30.54} & \textbf{31.24} &  \textbf{31.68} &  \textbf{31.71} \\
    \small{\lnw}      & \underline{29.34} & \underline{31.12} & \underline{32.35} & \underline{33.78} & 34.10 &  \underline{34.88} &  \underline{35.21} & \textbf{26.67} & \textbf{28.61} & \textbf{29.83} & \underline{30.21} & \underline{31.04} &  \underline{31.51} &  \underline{31.61} \\
    \midrule
     & \multicolumn{7}{c}{Ancestral Sampling} & \multicolumn{7}{c}{Ancestral Sampling} \\
    \cmidrule(l){2-8} \cmidrule(l){9-15}
    $|\mathcal{H}|$ & 4 &  8 & 16 & 32 & 64 & 128 & 256 & 4 &  8 & 16 & 32 & 64 & 128 & 256 \\
    \midrule
    \small{\baseline} & \textbf{24.31} & \textbf{25.74} & \textbf{27.75} & \underline{29.11} & \textbf{31.01} &  31.85 &  32.77 & \textbf{21.34} & \textbf{22.35} & \textbf{24.60} & \textbf{25.83} & \textbf{26.67} &  \underline{27.39} &  \underline{27.30} \\
    \small{\weighted} & \underline{23.84} & 25.48 & 27.43 & 29.04 & 30.70 &  \textbf{32.05} &  \underline{32.89} & 20.46 & 21.42 & 24.01 & 25.07 & 25.96 &  26.75 &  27.18 \\
    \small{\lnw}      & 23.78 & \underline{25.68} & \underline{27.71} & \textbf{29.19} & \underline{30.81} &  \underline{31.92} &  \textbf{33.05} & \underline{20.72} & \underline{21.88} & \underline{24.15} & \underline{25.44} & \underline{26.59} &  \textbf{27.49} &  \textbf{27.69}\\
    \midrule \midrule
    & \multicolumn{6}{l}{\#wins \lnw{} $>$ \baseline{}} & 24/28 & \multicolumn{6}{l}{\#wins \lnw{} $>$ \baseline{}} & 23/28 \\
    & \multicolumn{6}{l}{\#wins \weighted{} $>$ \baseline{}} & 23/28 & \multicolumn{6}{l}{\#wins \weighted{} $>$ \baseline{}} & 20/28 \\
    & \multicolumn{6}{l}{\#wins \lnw{} $>$ \weighted{}} & 7/28 & \multicolumn{6}{l}{\#wins \lnw{} $>$ \weighted{}} & 16/28 \\
    \bottomrule
    \end{tabular}
    }
    \caption{BLEU score of \weighted{} and \lnw{} with different sampling strategies on WMT'19 De-En and Ru-En. We evaluate using the first 1000 inputs of the dataset. The best score with the same sampling algorithm and number of samples is in bold. The second best score is underlined. Among the 56 settings (2 datasets, 4 sampling algorithms, with $|\mathcal{H}| \in \{4, 8, 16, 32, 64, 128, 256\}$), the total \#wins \lnw{} $>$ \baseline{} is 47/56, \#wins \weighted{} $>$ \baseline{} is 43/56, and \#wins \lnw{} $>$ \weighted{} is 23/56.}
    \label{tab:sampling}
\end{table*}

We evaluate MBR, MBMBR, and \lnw{} on machine translation, text summarization, and image captioning tasks.
We compare them using various probabilistic sampling algorithms which we describe in Section~\ref{sec:samplings}.
We use BERTScore \cite{bert-score} as the utility function of MBR for all experiments.
We use Huggingface's Transformers library to run all experiments \cite{wolf-etal-2020-transformers}.
For reproducibility, all experiments are performed using publicly available pretrained models and datasets.
Due to computational resource constraints, we evaluate the first 1000 inputs of each dataset. For SAMSum dataset (Section \ref{sec:sumsum}), we use the entire test dataset (819 inputs).

\subsection{Sampling Algorithms for MBR}
\label{sec:samplings}
The choice of sampling algorithm to collect the hypotheses has been studied extensively, as it has been shown to be critical to the performance of MBR.
While most of the classical work on MBR relies on beam search to generate samples, 
\citet{eikema-aziz-2020-map} propose to use unbiased sampling through ancestral sampling \cite{robert1999monte}.
\citet{fernandes-etal-2022-quality} have found that MBR with top-$k$ sampling \cite{fan-etal-2018-hierarchical} and nucleus sampling \cite{Holtzman2020The} generates much higher quality sequences compared to ancestral sampling.
Top-$k$ sampling is a simple modification of ancestral sampling that truncates all tokens except the $k$ most probable tokens. Similar to top-$k$ sampling, nucleus sampling also truncates the lower tail of the probability distribution. Nucleus sampling truncates all tokens except those in the {\it nucleus}, the smallest possible set of tokens that covers a fraction $p$ of the model probability.
\citet{freitag2023epsilon} show that epsilon sampling \cite{hewitt-etal-2022-truncation} further improves these sampling algorithms on machine translation tasks. Epsilon sampling is also a variant of ancestral sampling that truncates tokens whose probability is less than a fixed threshold $\epsilon$.
While the empirical distributions of top-$k$ sampling, nucleus sampling, and epsilon sampling are biased and inconsistent estimators of the model probability $\Pm$, MBR decoding using these sampling algorithms has been empirically shown to produce higher quality text than using ancestral sampling, which is unbiased and consistent. 

\subsection{Machine Translation}

We evaluate machine translation in two scenarios: (1) decoding a domain-specific machine translation model and (2) decoding an instruction-tuned language model with a prompt to trigger its translation capability.

\subsubsection{Machine Translation Model}
\label{sec:mtm}

We evaluate the performance of machine translation models using the WMT'19 dataset \cite{barrault-etal-2019-findings}.
WMT'19 dataset examines translation between English and other languages in the news domain.
We run experiments on four language pairs: English $\leftrightarrow$ German (En$\leftrightarrow$De) and English $\leftrightarrow$ Russian (En$\leftrightarrow$Ru) using the pretrained models of each language pair provided by fairseq \cite{ng-etal-2019-facebook}. 

We sample 256 sentences per source sentence for De and Ru$\rightarrow$En.
We use epsilon sampling \cite{hewitt-etal-2022-truncation,freitag2023epsilon}, top-$k$ sampling \cite{fan-etal-2018-hierarchical}, nucleus sampling \cite{Holtzman2020The}, and ancestral sampling. The parameters for the sampling methods are set according to the work of \citet{freitag2023epsilon}. For epsilon sampling, $\epsilon=0.02$. $k$ is set to $k=10$ for top-$k$ sampling. For nucleus sampling, $p=0.9$ is set. The temperature is set to $1.0$ for all algorithms.
We evaluate the BLEU score \cite{papineni-etal-2002-bleu} using the sacreBLEU library \cite{post-2018-call}. 

Table \ref{tab:sampling} shows the BLEU score from German and Russian to English.
Overall, we observe that \weighted{} and \lnw{} outperform MBR, except for ancestral sampling. As a reference, the BLEU score of a beam search with reranking is 40.8 for De-En and 40.0 for Ru-En \cite{ng-etal-2019-facebook}.
Table \ref{tab:wmt19en-x} shows the BLEU score of En-De and En-Ru with epsilon sampling. We observe that the proposed method outperforms the baseline in all settings.

We additionally evaluate the performance of MBMBR and MBR using utility functions trained for machine translation which we describe in Appendix \ref{sec:comet}.

\begin{table}
    \centering
    \adjustbox{max width=\columnwidth}{
    \begin{tabular}{lrrrrr}
    \toprule
    & \multicolumn{5}{c}{WMT'19 En-De} \\
    \cmidrule(l){2-6}
    $|\mathcal{H}|$ & 4 &  8 & 16 & 32 & 64  \\
    \midrule
    \small{\baseline}     & 29.03 & 30.73 & 32.10 & \underline{33.18} & \underline{34.04} \\
    \small{\weighted}     & \textbf{30.00} & \textbf{31.45} & \textbf{32.32} & 33.06 & 34.00  \\
    \small{\lnw} & \underline{29.94} & \underline{31.17} & \underline{32.26} & \textbf{33.35} & \textbf{34.41} \\\midrule\midrule
    & \multicolumn{5}{c}{WMT'19 En-Ru} \\
    \cmidrule(l){2-6}
    $|\mathcal{H}|$ & 4 &  8 & 16 & 32 & 64 \\\midrule
    \small{\baseline}     &  24.68 & 26.26 & \underline{27.25} & 27.80 & 27.97\\
    \small{\weighted}     & \textbf{25.52} & \textbf{26.72} & 27.19 & \textbf{28.14} & \textbf{28.53} \\
    \small{\lnw} &  \underline{25.15} & \underline{26.60} & \textbf{27.34} & \underline{28.08} & \underline{28.19} \\
    \bottomrule
    \end{tabular}
    }
    \caption{BLEU scores of MBMBR and MBR on WMT'19 En-De and En-Ru using machine translation models \cite{ng-etal-2019-facebook}. We use epsilon sampling with $\epsilon=0.02$. BERTScore is used as the utility function of MBR.
    Due to computational constraints, we evaluate using the first 1000 inputs of the dataset.}
    \label{tab:wmt19en-x}
\end{table}

\begin{table}
    \centering
    \adjustbox{max width=\columnwidth}{
    \begin{tabular}{lrrrrr}
    \toprule
    & \multicolumn{5}{c}{IWSLT'17 Fr-En} \\
    \cmidrule(l){2-6}
    $|\mathcal{H}|$ & 4 &  8 & 16 & 32 & 64  \\
    \midrule
    \small{\baseline}     & 25.33 & 26.89 & 27.94 & 28.48 & \underline{29.43} \\
    \small{\weighted}     &  \textbf{26.58} & \textbf{27.77} & \textbf{28.52} & \textbf{29.51} & 29.25 \\
    \small{\lnw} & \underline{26.26} & \underline{27.16} & \underline{28.38} & \underline{29.15} & \textbf{29.51} \\
    \bottomrule
    \end{tabular}
    }
    \caption{BLEU scores of MBMBR and MBR on IWSLT'17 using BLOOMZ and mT0 model \cite{muennighoff-etal-2023-crosslingual}. We use epsilon sampling with $\epsilon=0.02$. BERTScore is used as the utility function of MBR.
    Due to computational constraints, we evaluate using the first 1000 inputs of the dataset.}
    \label{tab:iwslt}
\end{table}

\subsubsection{Language Model with Prompting}
\label{sec:mt-llm}
For the experiments on a language model, we use IWSLT 2017 French $\rightarrow$ English dataset as a benchmark \cite{cettolo-etal-2017-overview}.
We use BLOOMZ and mT0 model (7.1B) loaded in 8-bit precision to reduce memory consumption \cite{muennighoff-etal-2023-crosslingual}.
64 sentences are sampled using epsilon sampling with $\epsilon=0.02$. The prompt is described in Appendix \ref{sec:prompt}.
The BLEU score of IWSLT'17 Fr-En is reported in Table \ref{tab:iwslt}.
We observe that MBMBR outperforms MBR, especially when the number of samples is small.
We additionally evaluate MBMBR using LLM-based machine translation models which we describe in Appendix \ref{sec:tower}.

\begin{table}
    \centering
    \adjustbox{max width=\columnwidth}{
    \begin{tabular}{lrrrrr}
    \toprule
    & \multicolumn{5}{c}{XSum} \\
    \cmidrule(l){2-6}
    $|\mathcal{H}|$ & 4 &  8 & 16 & 32 & 64  \\
    \midrule
    \small{\baseline}     & \underline{32.01} &  \underline{33.96} &   \underline{34.80} &   \underline{35.41} &   \underline{35.90} \\
    \small{\weighted}     & 31.69 &  33.09 &   34.00 &   34.72 &   35.51   \\
    \small{\lnw} &  \textbf{32.65} &  \textbf{34.01} &   \textbf{35.00} &   \textbf{35.63} &   \textbf{36.18} \\\midrule\midrule
    & \multicolumn{5}{c}{SAMSum} \\
    \cmidrule(l){2-6}
    $|\mathcal{H}|$ & 4 &  8 & 16 & 32 & 64 \\\midrule
    \small{\baseline}     & 30.55 &  31.16 &   31.98 &   32.51 &   33.03  \\
    \small{\weighted}     &  \textbf{31.68} &  \textbf{32.44} &   \textbf{33.14} &   \textbf{33.86} &   \textbf{34.50} \\
    \small{\lnw} &  \underline{31.17} &  \underline{32.06} &   \underline{32.62} &   \underline{32.99} &   \underline{33.14} \\
    \bottomrule
    \end{tabular}
    }
    \caption{ROUGE-L scores of MBMBR and MBR on XSum and SAMSum using text summarization models \cite{lewis-etal-2020-bart}. We use epsilon sampling with $\epsilon=0.02$. BERTScore is used as the utility function of MBR.
    Due to computational constraints, we evaluate using the first 1000 inputs of the dataset.}
    \label{tab:sumsum}
\end{table}

\begin{table}
    \centering
    \adjustbox{max width=\columnwidth}{
    \begin{tabular}{lrrrrr}
    \toprule
    & \multicolumn{5}{c}{CNN/DM} \\
    \cmidrule(l){2-6}
    $|\mathcal{H}|$ & 4 &  8 & 16 & 32 & 64  \\
    \midrule
    \small{\baseline}     & 16.72 &  16.62 &   16.94 &   17.55 &   17.26 \\
    \small{\weighted}     & \textbf{17.37} &  \textbf{17.16} &  \textbf{17.63} &   \textbf{18.07} &   \textbf{17.73} \\
    \small{\lnw} & \underline{16.75} &  \underline{16.75} &   \underline{16.98} &   \underline{17.63} &   \underline{17.38} \\
    \bottomrule
    \end{tabular}
    }
    \caption{ROUGE-L scores of MBMBR and MBR on CNN/DM using Mistral-7B-Instruct-v0.1 \cite{jiang2023mistral}. We use epsilon sampling with $\epsilon=0.02$. BERTScore is used as the utility function of MBR.
    Due to computational constraints, we evaluate using the first 1000 inputs of the dataset.}
    \label{tab:cnndm}
\end{table}

\subsection{Text Summarization}

We evaluate text summarization using (1) a BART model specifically fine-tuned for each text summarization dataset and (2) a general-purpose LLM with prompting.

\subsubsection{Text Summarization Model}
\label{sec:sumsum}

We use XSum \cite{narayan-etal-2018-dont} and SAMSum \cite{gliwa-etal-2019-samsum} datasets to evaluate the performance of MBMBR on the text summarization task.
XSum dataset is a benchmark for an abstractive single-document summarization. The documents are collected from BBC articles. 
We use a BART model pretrained on XSum dataset \cite{lewis-etal-2020-bart}.
SAMSum corpus is a collection of messenger-like conversations with summaries. We use a BART model pretrained on SAMSum dataset.
We evaluate ROUGE-L \cite{lin-2004-rouge} using HuggingFace's evaluate library. 

Although the performance of sampling algorithms has been extensively compared for machine translation tasks \cite{freitag2023epsilon}, little has been evaluated for other text generation tasks. As a preliminary experiment, we evaluate the performance of MBR using epsilon sampling ($\epsilon=0.02$), top-$k$ sampling ($k=10$), nucleus sampling ($p=0.9$), and ancestral sampling on text summarization tasks (XSum and SAMSum) and image captioning tasks (MS COCO and nocaps; see Section \ref{sec:captioning}). The results show that epsilon sampling outperforms others on all datasets (Appendix \ref{sec:eval-sampling}). Based on this observation, we will continue the evaluation of MBMBR and MBR using epsilon sampling with $\epsilon=0.02$.

The result using epsilon sampling is shown in Table \ref{tab:sumsum}. 
Overall, \lnw{} outperforms MBR in both datasets.
However, we observe that MBMBR drops in the ROUGE-L score in XSum dataset.
To investigate the effect of length normalization, we compute the relative length of the selected sentence compared to the reference text ($\ell(\vy) / \ell_{\mathrm{ref}}$) by the three methods. The average (standard deviation) of the relative length over the dataset of MBR, MBMBR, and \lnw{} are $0.870\ (\pm 0.268), 0.777\ (\pm 0.234)$, and $0.851\ (\pm 0.270)$, respectively. 
MBMBR has decreased the relative length by about $10$\% compared to MBR, which we speculate is the reason for the decrease in the ROUGE-L score. \lnw{} also becomes shorter but only by about $2$\%. 
For the SAMSum dataset, the decrease in relative length of MBMBR is about $2$\% which may be the reason why MBMBR successfully achieves a higher ROUGE-L score.

\subsubsection{Language Model with Prompting}
\label{sec:sum-llm}
CNN/DM is a collection of news articles and their summaries by journalists from CNN and the Daily Mail \cite{hermann2015teaching}.
We run experiments on CNN/DM with Mistral-7B-Instruct-v0.1 \cite{jiang2023mistral} loaded with 4-bit precision. See Appendix \ref{sec:prompt} for the prompt we use for CNN/DM.
The result is shown in Table \ref{tab:cnndm}. We see that MBMBR has a higher ROUGE-L score compared to MBR. 

\begin{table}
    \centering
    \adjustbox{max width=\columnwidth}{
    \begin{tabular}{lrrrrr}
    \toprule
    & \multicolumn{5}{c}{MSCOCO} \\
    \cmidrule(l){2-6}
    $|\mathcal{H}|$ & 4 &  8 & 16 & 32 & 64  \\
    \midrule
    \small{\baseline}     & 27.12 & 28.46 & 30.42 & 31.51 & 33.24  \\
    \small{\weighted}     & \textbf{28.75} & \textbf{30.56} & \textbf{33.12} & \textbf{34.25} & \textbf{34.96}  \\
    \small{\lnw} & \underline{28.65} & \underline{29.29} & \underline{31.14} & \underline{32.54} & \underline{33.99} \\\midrule\midrule
    & \multicolumn{5}{c}{nocaps} \\
    \cmidrule(l){2-6}
    $|\mathcal{H}|$ & 4 &  8 & 16 & 32 & 64 \\\midrule
    \small{\baseline}     & \underline{25.43} & \underline{27.67} & \underline{28.73} & \underline{30.93} & \underline{31.42} \\
    \small{\weighted}     &  23.31 & 24.73 & 25.87 & 26.91 & 27.27  \\
    \small{\lnw} &  \textbf{28.86} & \textbf{29.78} & \textbf{30.47} & \textbf{32.10} & \textbf{32.68} \\
    \bottomrule
    \end{tabular}
    }
    \caption{BLEU scores of MBMBR and MBR on MSCOCO and nocaps using BLIP-2 \cite{pmlr-v202-li23q} with Flan T5-xl \cite{chung2022scaling}. We use epsilon sampling with $\epsilon=0.02$. BERTScore is used as the utility function of MBR.
    Due to computational constraints, we evaluate using the first 1000 inputs of the dataset.}
    \label{tab:captioning}
\end{table}

\begin{table}
    \centering
    \adjustbox{max width=\columnwidth}{
    \begin{tabular}{lrrrrr}
    \toprule
    & \multicolumn{5}{c}{E2E NLG} \\
    \cmidrule(l){2-6}
    $|\mathcal{H}|$ & 4 &  8 & 16 & 32 & 64  \\
    \midrule
    \small{\baseline}     & \textbf{20.49} & \underline{20.77} & \underline{20.42} & \textbf{21.04} & \textbf{22.30} \\
    \small{\weighted}     & 17.79 & 17.01 & 15.80 & 16.26 & 16.07 \\
    \small{\lnw} &  \underline{19.93} & \textbf{21.28} & \textbf{20.88} & \underline{20.99} & \underline{22.03}  \\
    \bottomrule
    \end{tabular}
    }
    \caption{BLEU scores of MBMBR and MBR on E2E NLG using Mistral-7B-Instruct-v0.1 \cite{jiang2023mistral}. We use epsilon sampling with $\epsilon=0.02$. BERTScore is used as the utility function of MBR.
    Due to computational constraints, we evaluate using the first 1000 inputs of the dataset.}
    \label{tab:e2e}
\end{table}

\subsection{Image Captioning with BLIP-2}
\label{sec:captioning}

We evaluate our method for image captioning tasks. Although the input for the image captioning task is an image rather than a text, MBR is applicable as is. That is, we generate a set of hypotheses conditional on the input image and then select the best hypothesis using the MBR objective.

We use BLIP-2 \cite{pmlr-v202-li23q} with Flan T5-xl \cite{chung2022scaling} loaded in 8-bit to evaluate on two datasets: MS COCO dataset \cite{lin2014microsoft} and nocaps dataset \cite{agrawal2019nocaps}. We use a fine-tuned model for MS COCO and a base model for nocaps. 
Overall, \lnw{} outperforms MBR in both datasets (Table \ref{tab:captioning}). 


\subsection{Data-to-Text with Few-Shot Learning}
\label{sec:e2e}

We evaluate the proposed method on a data-to-text generation task using E2E NLG dataset \cite{novikova-etal-2017-e2e}. E2E NLG is a data-driven natural language generation task in the restaurant domain. 
Given a set of key-value pairs about a restaurant, the task is to provide a short English description of the restaurant in a few sentences.
We use a Mistral-7B-Instruct-v0.1 \cite{jiang2023mistral} loaded with 4-bit with a prompt provided by \citet{suzgun-etal-2023-follow}.
The result is shown in Table \ref{tab:e2e}. Overall, we observe no improvement over the baseline in E2E NLG dataset. We speculate that this is because the quality of the sentences generated is not high enough for the MBR to work properly.

\section{Related Work}

MBR decoding has been studied in many NLP tasks including parsing \cite{goodman-1996-parsing}, speech recognition \cite{goel2000minimum}, bilingual word alignment \cite{kumar-byrne-2002-minimum}, and machine translation \cite{kumar-byrne-2004-minimum}.
MBR decoding has recently gained attention in machine translation to overcome some of the biases of MAP decoding in neural machine translation \cite{eikema-aziz-2020-map,muller-sennrich-2021-understanding,eikema-aziz-2022-sampling}. 

\citet{freitag-etal-2022-high} and \citet{fernandes-etal-2022-quality} show that using neural-based utility functions such as BLEURT \cite{sellam-etal-2020-bleurt,pu-etal-2021-learning} and COMET \cite{rei-etal-2020-unbabels,rei-etal-2022-comet} instead of lexical overlap metrics (e.g., BLEU) further improves the output quality of the MBR decoding.
The improvement in utility function is orthogonal to our approach as our improvement is on the estimation of the distribution.


The main drawback of MBR decoding is that it is computationally intensive. 
\citet{cheng-vlachos-2023-faster} shows that the number of calls to the utility functions can be significantly reduced by iteratively reducing the number of candidates.
\citet{finkelstein2023mbr} and \citet{yang2023direct} show that self-training a machine translation model using its own MBR-decoded output can improve the performance of more efficient decoding methods such as beam search.


\section{Conclusion}

We propose model-based MBR (MBMBR) decoding, a variant of MBR that uses a model-based distribution as an estimator of the model probability. 
We evaluate the model-based distribution analytically and empirically to show that it is closer to the true model probability than the Monte Carlo estimate with respect to KL divergence. The result suggests that the model-based distribution is likely to be a better estimator for the purpose of MBR decoding.
We perform MBMBR decoding on a variety of text generation tasks, including machine translation, text summarization, image captioning, and data-to-text, using both domain-specific sequence-to-sequence models and domain-independent large language models.
The empirical results show that MBMBR outperforms MBR in most cases. 
We believe that MBMBR will be a practical choice for future MBR decoding because of its applicability and significant performance improvements.

\section*{Acknowledgments}

We thank all the reviewers for their constructive comments throughout the manuscript review cycles.
We thank William Byrne for the insightful feedback and for improving the proof of Theorem 4.1. Kaito Ariu's research is supported by JSPS KAKENHI Grant No. 23K19986. 

\section*{Impact Statement}

While language generation can be used for malicious purposes, we do not foresee any specific ethical concerns or societal implications with the analysis in this paper beyond those discussed by \citet{Bender2021}.


\begin{thebibliography}{62}
\providecommand{\natexlab}[1]{#1}
\providecommand{\url}[1]{\texttt{#1}}
\expandafter\ifx\csname urlstyle\endcsname\relax
  \providecommand{\doi}[1]{doi: #1}\else
  \providecommand{\doi}{doi: \begingroup \urlstyle{rm}\Url}\fi

\bibitem[Agrawal et~al.(2019)Agrawal, Anderson, Desai, Wang, Chen, Jain, Johnson, Batra, Parikh, and Lee]{agrawal2019nocaps}
Agrawal, H., Anderson, P., Desai, K., Wang, Y., Chen, X., Jain, R., Johnson, M., Batra, D., Parikh, D., and Lee, S.
\newblock nocaps: novel object captioning at scale.
\newblock In \emph{2019 {IEEE/CVF} International Conference on Computer Vision, {ICCV} 2019, Seoul, Korea (South)}, pp.\  8947--8956. {IEEE}, 2019.
\newblock \doi{10.1109/ICCV.2019.00904}.
\newblock URL \url{https://doi.org/10.1109/ICCV.2019.00904}.

\bibitem[Alves et~al.(2024)Alves, Pombal, Guerreiro, Martins, Alves, Farajian, Peters, Rei, Fernandes, Agrawal, Colombo, de~Souza, and Martins]{alves2024tower}
Alves, D.~M., Pombal, J., Guerreiro, N.~M., Martins, P.~H., Alves, J., Farajian, A., Peters, B., Rei, R., Fernandes, P., Agrawal, S., Colombo, P., de~Souza, J. G.~C., and Martins, A. F.~T.
\newblock Tower: An open multilingual large language model for translation-related tasks.
\newblock \emph{arXiv preprint arXiv:2402.17733}, 2024.

\bibitem[Anderson et~al.(2017)Anderson, Fernando, Johnson, and Gould]{anderson-etal-2017-guided}
Anderson, P., Fernando, B., Johnson, M., and Gould, S.
\newblock Guided open vocabulary image captioning with constrained beam search.
\newblock In \emph{Proceedings of the 2017 Conference on Empirical Methods in Natural Language Processing}, pp.\  936--945, Copenhagen, Denmark, September 2017. Association for Computational Linguistics.
\newblock \doi{10.18653/v1/D17-1098}.
\newblock URL \url{https://aclanthology.org/D17-1098}.

\bibitem[Barrault et~al.(2019)Barrault, Bojar, Costa-juss{\`a}, Federmann, Fishel, Graham, Haddow, Huck, Koehn, Malmasi, Monz, M{\"u}ller, Pal, Post, and Zampieri]{barrault-etal-2019-findings}
Barrault, L., Bojar, O., Costa-juss{\`a}, M.~R., Federmann, C., Fishel, M., Graham, Y., Haddow, B., Huck, M., Koehn, P., Malmasi, S., Monz, C., M{\"u}ller, M., Pal, S., Post, M., and Zampieri, M.
\newblock Findings of the 2019 conference on machine translation ({WMT}19).
\newblock In \emph{Proceedings of the Fourth Conference on Machine Translation (Volume 2: Shared Task Papers, Day 1)}, pp.\  1--61, Florence, Italy, August 2019. Association for Computational Linguistics.
\newblock \doi{10.18653/v1/W19-5301}.
\newblock URL \url{https://aclanthology.org/W19-5301}.

\bibitem[Bender et~al.(2021)Bender, Gebru, McMillan-Major, and Shmitchell]{Bender2021}
Bender, E.~M., Gebru, T., McMillan-Major, A., and Shmitchell, S.
\newblock On the dangers of stochastic parrots: Can language models be too big?
\newblock In \emph{Proceedings of the 2021 ACM Conference on Fairness, Accountability, and Transparency}, FAccT '21, pp.\  610–623, New York, NY, USA, 2021. Association for Computing Machinery.
\newblock ISBN 9781450383097.
\newblock \doi{10.1145/3442188.3445922}.
\newblock URL \url{https://doi.org/10.1145/3442188.3445922}.

\bibitem[Bertsch et~al.(2023)Bertsch, Xie, Neubig, and Gormley]{bertschs2023}
Bertsch, A., Xie, A., Neubig, G., and Gormley, M.
\newblock It{'}s {MBR} all the way down: Modern generation techniques through the lens of minimum {B}ayes risk.
\newblock In Elazar, Y., Ettinger, A., Kassner, N., Ruder, S., and A.~Smith, N. (eds.), \emph{Proceedings of the Big Picture Workshop}, pp.\  108--122, Singapore, Singapore, December 2023. Association for Computational Linguistics.
\newblock URL \url{https://aclanthology.org/2023.bigpicture-1.9}.

\bibitem[Cettolo et~al.(2017)Cettolo, Federico, Bentivogli, Niehues, St{\"u}ker, Sudoh, Yoshino, and Federmann]{cettolo-etal-2017-overview}
Cettolo, M., Federico, M., Bentivogli, L., Niehues, J., St{\"u}ker, S., Sudoh, K., Yoshino, K., and Federmann, C.
\newblock Overview of the {IWSLT} 2017 evaluation campaign.
\newblock In \emph{Proceedings of the 14th International Conference on Spoken Language Translation}, pp.\  2--14, Tokyo, Japan, December 14-15 2017. International Workshop on Spoken Language Translation.
\newblock URL \url{https://aclanthology.org/2017.iwslt-1.1}.

\bibitem[Cheng \& Vlachos(2023)Cheng and Vlachos]{cheng-vlachos-2023-faster}
Cheng, J. and Vlachos, A.
\newblock Faster minimum {B}ayes risk decoding with confidence-based pruning.
\newblock In Bouamor, H., Pino, J., and Bali, K. (eds.), \emph{Proceedings of the 2023 Conference on Empirical Methods in Natural Language Processing}, pp.\  12473--12480, Singapore, December 2023. Association for Computational Linguistics.
\newblock URL \url{https://aclanthology.org/2023.emnlp-main.767}.

\bibitem[Chung et~al.(2022)Chung, Hou, Longpre, Zoph, Tay, Fedus, Li, Wang, Dehghani, Brahma, et~al.]{chung2022scaling}
Chung, H.~W., Hou, L., Longpre, S., Zoph, B., Tay, Y., Fedus, W., Li, E., Wang, X., Dehghani, M., Brahma, S., et~al.
\newblock Scaling instruction-finetuned language models.
\newblock \emph{arXiv preprint arXiv:2210.11416}, 2022.

\bibitem[Cohen \& Beck(2019)Cohen and Beck]{pmlr-v97-cohen19a}
Cohen, E. and Beck, C.
\newblock Empirical analysis of beam search performance degradation in neural sequence models.
\newblock In Chaudhuri, K. and Salakhutdinov, R. (eds.), \emph{Proceedings of the 36th International Conference on Machine Learning}, volume~97 of \emph{Proceedings of Machine Learning Research}, pp.\  1290--1299. PMLR, 09--15 Jun 2019.
\newblock URL \url{https://proceedings.mlr.press/v97/cohen19a.html}.

\bibitem[Cover \& Thomas(2006)Cover and Thomas]{cover1999elements}
Cover, T.~M. and Thomas, J.~A.
\newblock \emph{Elements of Information Theory}.
\newblock Wiley-Interscience, USA, 2006.
\newblock ISBN 0471241954.

\bibitem[Csisz{\'{a}}r \& K{\"{o}}rner(2011)Csisz{\'{a}}r and K{\"{o}}rner]{DBLP:books/cu/CsiszarK11}
Csisz{\'{a}}r, I. and K{\"{o}}rner, J.
\newblock \emph{Information Theory - Coding Theorems for Discrete Memoryless Systems, Second Edition}.
\newblock Cambridge University Press, 2011.
\newblock ISBN 978-0-51192188-9.
\newblock \doi{10.1017/CBO9780511921889}.
\newblock URL \url{https://doi.org/10.1017/CBO9780511921889}.

\bibitem[Eikema \& Aziz(2020)Eikema and Aziz]{eikema-aziz-2020-map}
Eikema, B. and Aziz, W.
\newblock Is {MAP} decoding all you need? the inadequacy of the mode in neural machine translation.
\newblock In \emph{Proceedings of the 28th International Conference on Computational Linguistics}, pp.\  4506--4520, Barcelona, Spain (Online), December 2020. International Committee on Computational Linguistics.
\newblock \doi{10.18653/v1/2020.coling-main.398}.
\newblock URL \url{https://aclanthology.org/2020.coling-main.398}.

\bibitem[Eikema \& Aziz(2022)Eikema and Aziz]{eikema-aziz-2022-sampling}
Eikema, B. and Aziz, W.
\newblock Sampling-based approximations to minimum {B}ayes risk decoding for neural machine translation.
\newblock In \emph{Proceedings of the 2022 Conference on Empirical Methods in Natural Language Processing}, pp.\  10978--10993, Abu Dhabi, United Arab Emirates, December 2022. Association for Computational Linguistics.
\newblock \doi{10.18653/v1/2022.emnlp-main.754}.
\newblock URL \url{https://aclanthology.org/2022.emnlp-main.754}.

\bibitem[Fan et~al.(2018)Fan, Lewis, and Dauphin]{fan-etal-2018-hierarchical}
Fan, A., Lewis, M., and Dauphin, Y.
\newblock Hierarchical neural story generation.
\newblock In \emph{Proceedings of the 56th Annual Meeting of the Association for Computational Linguistics (Volume 1: Long Papers)}, pp.\  889--898, Melbourne, Australia, July 2018. Association for Computational Linguistics.
\newblock \doi{10.18653/v1/P18-1082}.
\newblock URL \url{https://aclanthology.org/P18-1082}.

\bibitem[Farinhas et~al.(2023)Farinhas, de~Souza, and Martins]{farinhas2023empirical}
Farinhas, A., de~Souza, J., and Martins, A.
\newblock An empirical study of translation hypothesis ensembling with large language models.
\newblock In Bouamor, H., Pino, J., and Bali, K. (eds.), \emph{Proceedings of the 2023 Conference on Empirical Methods in Natural Language Processing}, pp.\  11956--11970, Singapore, December 2023. Association for Computational Linguistics.
\newblock \doi{10.18653/v1/2023.emnlp-main.733}.
\newblock URL \url{https://aclanthology.org/2023.emnlp-main.733}.

\bibitem[Fernandes et~al.(2022)Fernandes, Farinhas, Rei, C.~de Souza, Ogayo, Neubig, and Martins]{fernandes-etal-2022-quality}
Fernandes, P., Farinhas, A., Rei, R., C.~de Souza, J.~G., Ogayo, P., Neubig, G., and Martins, A.
\newblock Quality-aware decoding for neural machine translation.
\newblock In \emph{Proceedings of the 2022 Conference of the North American Chapter of the Association for Computational Linguistics: Human Language Technologies}, pp.\  1396--1412, Seattle, United States, July 2022. Association for Computational Linguistics.
\newblock \doi{10.18653/v1/2022.naacl-main.100}.
\newblock URL \url{https://aclanthology.org/2022.naacl-main.100}.

\bibitem[Finkelstein \& Freitag(2024)Finkelstein and Freitag]{finkelstein2023mbr}
Finkelstein, M. and Freitag, M.
\newblock {MBR} and {QE} finetuning: Training-time distillation of the best and most expensive decoding methods.
\newblock In \emph{The Twelfth International Conference on Learning Representations}, 2024.
\newblock URL \url{https://openreview.net/forum?id=bkNx3O0sND}.

\bibitem[Freitag et~al.(2022)Freitag, Grangier, Tan, and Liang]{freitag-etal-2022-high}
Freitag, M., Grangier, D., Tan, Q., and Liang, B.
\newblock High quality rather than high model probability: Minimum {B}ayes risk decoding with neural metrics.
\newblock \emph{Transactions of the Association for Computational Linguistics}, 10:\penalty0 811--825, 2022.
\newblock \doi{10.1162/tacl_a_00491}.
\newblock URL \url{https://aclanthology.org/2022.tacl-1.47}.

\bibitem[Freitag et~al.(2023)Freitag, Ghorbani, and Fernandes]{freitag2023epsilon}
Freitag, M., Ghorbani, B., and Fernandes, P.
\newblock Epsilon sampling rocks: Investigating sampling strategies for minimum {B}ayes risk decoding for machine translation.
\newblock In Bouamor, H., Pino, J., and Bali, K. (eds.), \emph{Findings of the Association for Computational Linguistics: EMNLP 2023}, pp.\  9198--9209, Singapore, December 2023. Association for Computational Linguistics.
\newblock URL \url{https://aclanthology.org/2023.findings-emnlp.617}.

\bibitem[Gliwa et~al.(2019)Gliwa, Mochol, Biesek, and Wawer]{gliwa-etal-2019-samsum}
Gliwa, B., Mochol, I., Biesek, M., and Wawer, A.
\newblock {SAMS}um corpus: A human-annotated dialogue dataset for abstractive summarization.
\newblock In \emph{Proceedings of the 2nd Workshop on New Frontiers in Summarization}, pp.\  70--79, Hong Kong, China, November 2019. Association for Computational Linguistics.
\newblock \doi{10.18653/v1/D19-5409}.
\newblock URL \url{https://aclanthology.org/D19-5409}.

\bibitem[Goel \& Byrne(2000)Goel and Byrne]{goel2000minimum}
Goel, V. and Byrne, W.~J.
\newblock Minimum bayes-risk automatic speech recognition.
\newblock \emph{Computer Speech \& Language}, 14\penalty0 (2):\penalty0 115--135, 2000.

\bibitem[Goodman(1996)]{goodman-1996-parsing}
Goodman, J.
\newblock Parsing algorithms and metrics.
\newblock In \emph{34th Annual Meeting of the Association for Computational Linguistics}, pp.\  177--183, Santa Cruz, California, USA, June 1996. Association for Computational Linguistics.
\newblock \doi{10.3115/981863.981887}.
\newblock URL \url{https://aclanthology.org/P96-1024}.

\bibitem[Graves(2012)]{graves2012sequence}
Graves, A.
\newblock Sequence transduction with recurrent neural networks.
\newblock \emph{arXiv preprint arXiv:1211.3711}, 2012.

\bibitem[Hermann et~al.(2015)Hermann, Kocisk{\'{y}}, Grefenstette, Espeholt, Kay, Suleyman, and Blunsom]{hermann2015teaching}
Hermann, K.~M., Kocisk{\'{y}}, T., Grefenstette, E., Espeholt, L., Kay, W., Suleyman, M., and Blunsom, P.
\newblock Teaching machines to read and comprehend.
\newblock In Cortes, C., Lawrence, N.~D., Lee, D.~D., Sugiyama, M., and Garnett, R. (eds.), \emph{Advances in Neural Information Processing Systems 28: Annual Conference on Neural Information Processing Systems 2015, December 7-12, 2015, Montreal, Quebec, Canada}, pp.\  1693--1701, 2015.
\newblock URL \url{https://proceedings.neurips.cc/paper/2015/hash/afdec7005cc9f14302cd0474fd0f3c96-Abstract.html}.

\bibitem[Hewitt et~al.(2022)Hewitt, Manning, and Liang]{hewitt-etal-2022-truncation}
Hewitt, J., Manning, C., and Liang, P.
\newblock Truncation sampling as language model desmoothing.
\newblock In \emph{Findings of the Association for Computational Linguistics: EMNLP 2022}, pp.\  3414--3427, Abu Dhabi, United Arab Emirates, December 2022. Association for Computational Linguistics.
\newblock \doi{10.18653/v1/2022.findings-emnlp.249}.
\newblock URL \url{https://aclanthology.org/2022.findings-emnlp.249}.

\bibitem[Holtzman et~al.(2020)Holtzman, Buys, Du, Forbes, and Choi]{Holtzman2020The}
Holtzman, A., Buys, J., Du, L., Forbes, M., and Choi, Y.
\newblock The curious case of neural text degeneration.
\newblock In \emph{8th International Conference on Learning Representations, {ICLR} 2020, Addis Ababa, Ethiopia, April 26-30, 2020}. OpenReview.net, 2020.
\newblock URL \url{https://openreview.net/forum?id=rygGQyrFvH}.

\bibitem[Jiang et~al.(2023)Jiang, Sablayrolles, Mensch, Bamford, Chaplot, de~las Casas, Bressand, Lengyel, Lample, Saulnier, Lavaud, Lachaux, Stock, Scao, Lavril, Wang, Lacroix, and Sayed]{jiang2023mistral}
Jiang, A.~Q., Sablayrolles, A., Mensch, A., Bamford, C., Chaplot, D.~S., de~las Casas, D., Bressand, F., Lengyel, G., Lample, G., Saulnier, L., Lavaud, L.~R., Lachaux, M.-A., Stock, P., Scao, T.~L., Lavril, T., Wang, T., Lacroix, T., and Sayed, W.~E.
\newblock Mistral 7b.
\newblock \emph{arXiv preprint arxiv:2310.06825}, 2023.

\bibitem[Koehn \& Knowles(2017)Koehn and Knowles]{koehn-knowles-2017-six}
Koehn, P. and Knowles, R.
\newblock Six challenges for neural machine translation.
\newblock In \emph{Proceedings of the First Workshop on Neural Machine Translation}, pp.\  28--39, Vancouver, August 2017. Association for Computational Linguistics.
\newblock \doi{10.18653/v1/W17-3204}.
\newblock URL \url{https://aclanthology.org/W17-3204}.

\bibitem[Kumar \& Byrne(2002)Kumar and Byrne]{kumar-byrne-2002-minimum}
Kumar, S. and Byrne, W.
\newblock Minimum {B}ayes-risk word alignments of bilingual texts.
\newblock In \emph{Proceedings of the 2002 Conference on Empirical Methods in Natural Language Processing ({EMNLP} 2002)}, pp.\  140--147. Association for Computational Linguistics, July 2002.
\newblock \doi{10.3115/1118693.1118712}.
\newblock URL \url{https://aclanthology.org/W02-1019}.

\bibitem[Kumar \& Byrne(2004)Kumar and Byrne]{kumar-byrne-2004-minimum}
Kumar, S. and Byrne, W.
\newblock Minimum {B}ayes-risk decoding for statistical machine translation.
\newblock In \emph{Proceedings of the Human Language Technology Conference of the North {A}merican Chapter of the Association for Computational Linguistics: {HLT}-{NAACL} 2004}, pp.\  169--176, Boston, Massachusetts, USA, May 2 - May 7 2004. Association for Computational Linguistics.
\newblock URL \url{https://aclanthology.org/N04-1022}.

\bibitem[Lewis et~al.(2020)Lewis, Liu, Goyal, Ghazvininejad, Mohamed, Levy, Stoyanov, and Zettlemoyer]{lewis-etal-2020-bart}
Lewis, M., Liu, Y., Goyal, N., Ghazvininejad, M., Mohamed, A., Levy, O., Stoyanov, V., and Zettlemoyer, L.
\newblock {BART}: Denoising sequence-to-sequence pre-training for natural language generation, translation, and comprehension.
\newblock In \emph{Proceedings of the 58th Annual Meeting of the Association for Computational Linguistics}, pp.\  7871--7880, Online, July 2020. Association for Computational Linguistics.
\newblock \doi{10.18653/v1/2020.acl-main.703}.
\newblock URL \url{https://aclanthology.org/2020.acl-main.703}.

\bibitem[Li et~al.(2023)Li, Li, Savarese, and Hoi]{pmlr-v202-li23q}
Li, J., Li, D., Savarese, S., and Hoi, S.
\newblock {BLIP}-2: Bootstrapping language-image pre-training with frozen image encoders and large language models.
\newblock In Krause, A., Brunskill, E., Cho, K., Engelhardt, B., Sabato, S., and Scarlett, J. (eds.), \emph{Proceedings of the 40th International Conference on Machine Learning}, volume 202 of \emph{Proceedings of Machine Learning Research}, pp.\  19730--19742. PMLR, 23--29 Jul 2023.
\newblock URL \url{https://proceedings.mlr.press/v202/li23q.html}.

\bibitem[Lin(2004)]{lin-2004-rouge}
Lin, C.-Y.
\newblock {ROUGE}: A package for automatic evaluation of summaries.
\newblock In \emph{Text Summarization Branches Out}, pp.\  74--81, Barcelona, Spain, July 2004. Association for Computational Linguistics.
\newblock URL \url{https://aclanthology.org/W04-1013}.

\bibitem[Lin et~al.(2014)Lin, Maire, Belongie, Hays, Perona, Ramanan, Doll{\'a}r, and Zitnick]{lin2014microsoft}
Lin, T.-Y., Maire, M., Belongie, S., Hays, J., Perona, P., Ramanan, D., Doll{\'a}r, P., and Zitnick, C.~L.
\newblock Microsoft {COCO}: Common objects in context.
\newblock In \emph{Computer Vision--ECCV 2014: 13th European Conference, Zurich, Switzerland, September 6-12, 2014, Proceedings, Part V 13}, pp.\  740--755. Springer, 2014.

\bibitem[Meister et~al.(2020)Meister, Cotterell, and Vieira]{meister-etal-2020-beam}
Meister, C., Cotterell, R., and Vieira, T.
\newblock If beam search is the answer, what was the question?
\newblock In \emph{Proceedings of the 2020 Conference on Empirical Methods in Natural Language Processing (EMNLP)}, pp.\  2173--2185, Online, November 2020. Association for Computational Linguistics.
\newblock \doi{10.18653/v1/2020.emnlp-main.170}.
\newblock URL \url{https://aclanthology.org/2020.emnlp-main.170}.

\bibitem[Muennighoff et~al.(2023)Muennighoff, Wang, Sutawika, Roberts, Biderman, Le~Scao, Bari, Shen, Yong, Schoelkopf, Tang, Radev, Aji, Almubarak, Albanie, Alyafeai, Webson, Raff, and Raffel]{muennighoff-etal-2023-crosslingual}
Muennighoff, N., Wang, T., Sutawika, L., Roberts, A., Biderman, S., Le~Scao, T., Bari, M.~S., Shen, S., Yong, Z.~X., Schoelkopf, H., Tang, X., Radev, D., Aji, A.~F., Almubarak, K., Albanie, S., Alyafeai, Z., Webson, A., Raff, E., and Raffel, C.
\newblock Crosslingual generalization through multitask finetuning.
\newblock In \emph{Proceedings of the 61st Annual Meeting of the Association for Computational Linguistics (Volume 1: Long Papers)}, pp.\  15991--16111, Toronto, Canada, July 2023. Association for Computational Linguistics.
\newblock \doi{10.18653/v1/2023.acl-long.891}.
\newblock URL \url{https://aclanthology.org/2023.acl-long.891}.

\bibitem[M{\"u}ller \& Sennrich(2021)M{\"u}ller and Sennrich]{muller-sennrich-2021-understanding}
M{\"u}ller, M. and Sennrich, R.
\newblock Understanding the properties of minimum {B}ayes risk decoding in neural machine translation.
\newblock In \emph{Proceedings of the 59th Annual Meeting of the Association for Computational Linguistics and the 11th International Joint Conference on Natural Language Processing (Volume 1: Long Papers)}, pp.\  259--272, Online, August 2021. Association for Computational Linguistics.
\newblock \doi{10.18653/v1/2021.acl-long.22}.
\newblock URL \url{https://aclanthology.org/2021.acl-long.22}.

\bibitem[Murray \& Chiang(2018)Murray and Chiang]{murray-chiang-2018-correcting}
Murray, K. and Chiang, D.
\newblock Correcting length bias in neural machine translation.
\newblock In \emph{Proceedings of the Third Conference on Machine Translation: Research Papers}, pp.\  212--223, Brussels, Belgium, October 2018. Association for Computational Linguistics.
\newblock \doi{10.18653/v1/W18-6322}.
\newblock URL \url{https://aclanthology.org/W18-6322}.

\bibitem[Narayan et~al.(2018)Narayan, Cohen, and Lapata]{narayan-etal-2018-dont}
Narayan, S., Cohen, S.~B., and Lapata, M.
\newblock Don{'}t give me the details, just the summary! topic-aware convolutional neural networks for extreme summarization.
\newblock In \emph{Proceedings of the 2018 Conference on Empirical Methods in Natural Language Processing}, pp.\  1797--1807, Brussels, Belgium, October-November 2018. Association for Computational Linguistics.
\newblock \doi{10.18653/v1/D18-1206}.
\newblock URL \url{https://aclanthology.org/D18-1206}.

\bibitem[Ng et~al.(2019)Ng, Yee, Baevski, Ott, Auli, and Edunov]{ng-etal-2019-facebook}
Ng, N., Yee, K., Baevski, A., Ott, M., Auli, M., and Edunov, S.
\newblock {F}acebook {FAIR}{'}s {WMT}19 news translation task submission.
\newblock In \emph{Proceedings of the Fourth Conference on Machine Translation (Volume 2: Shared Task Papers, Day 1)}, pp.\  314--319, Florence, Italy, August 2019. Association for Computational Linguistics.
\newblock \doi{10.18653/v1/W19-5333}.
\newblock URL \url{https://aclanthology.org/W19-5333}.

\bibitem[Novikova et~al.(2017)Novikova, Du{\v{s}}ek, and Rieser]{novikova-etal-2017-e2e}
Novikova, J., Du{\v{s}}ek, O., and Rieser, V.
\newblock The {E}2{E} dataset: New challenges for end-to-end generation.
\newblock In \emph{Proceedings of the 18th Annual {SIG}dial Meeting on Discourse and Dialogue}, pp.\  201--206, Saarbr{\"u}cken, Germany, August 2017. Association for Computational Linguistics.
\newblock \doi{10.18653/v1/W17-5525}.
\newblock URL \url{https://aclanthology.org/W17-5525}.

\bibitem[Ohashi et~al.(2024)Ohashi, Honda, Morimura, and Jinnai]{ohashi2024true}
Ohashi, A., Honda, U., Morimura, T., and Jinnai, Y.
\newblock On the true distribution approximation of minimum bayes-risk decoding.
\newblock In \emph{Proceedings of the 2024 Conference of the North American Chapter of the Association for Computational Linguistics: Human Language Technologies}, Mexico City, Mexico, July 2024. Association for Computational Linguistics.

\bibitem[Ott et~al.(2019)Ott, Edunov, Baevski, Fan, Gross, Ng, Grangier, and Auli]{ott-etal-2019-fairseq}
Ott, M., Edunov, S., Baevski, A., Fan, A., Gross, S., Ng, N., Grangier, D., and Auli, M.
\newblock fairseq: A fast, extensible toolkit for sequence modeling.
\newblock In \emph{Proceedings of the 2019 Conference of the North {A}merican Chapter of the Association for Computational Linguistics (Demonstrations)}, pp.\  48--53, Minneapolis, Minnesota, June 2019. Association for Computational Linguistics.
\newblock \doi{10.18653/v1/N19-4009}.
\newblock URL \url{https://aclanthology.org/N19-4009}.

\bibitem[Papineni et~al.(2002)Papineni, Roukos, Ward, and Zhu]{papineni-etal-2002-bleu}
Papineni, K., Roukos, S., Ward, T., and Zhu, W.-J.
\newblock {B}leu: a method for automatic evaluation of machine translation.
\newblock In \emph{Proceedings of the 40th Annual Meeting of the Association for Computational Linguistics}, pp.\  311--318, Philadelphia, Pennsylvania, USA, July 2002. Association for Computational Linguistics.
\newblock \doi{10.3115/1073083.1073135}.
\newblock URL \url{https://aclanthology.org/P02-1040}.

\bibitem[Piantadosi(2014)]{piantadosi2014zipf}
Piantadosi, S.~T.
\newblock Zipf’s word frequency law in natural language: A critical review and future directions.
\newblock \emph{Psychonomic bulletin \& review}, 21:\penalty0 1112--1130, 2014.

\bibitem[Post(2018)]{post-2018-call}
Post, M.
\newblock A call for clarity in reporting {BLEU} scores.
\newblock In \emph{Proceedings of the Third Conference on Machine Translation: Research Papers}, pp.\  186--191, Brussels, Belgium, October 2018. Association for Computational Linguistics.
\newblock \doi{10.18653/v1/W18-6319}.
\newblock URL \url{https://aclanthology.org/W18-6319}.

\bibitem[Pu et~al.(2021)Pu, Chung, Parikh, Gehrmann, and Sellam]{pu-etal-2021-learning}
Pu, A., Chung, H.~W., Parikh, A., Gehrmann, S., and Sellam, T.
\newblock Learning compact metrics for {MT}.
\newblock In \emph{Proceedings of the 2021 Conference on Empirical Methods in Natural Language Processing}, pp.\  751--762, Online and Punta Cana, Dominican Republic, November 2021. Association for Computational Linguistics.
\newblock \doi{10.18653/v1/2021.emnlp-main.58}.
\newblock URL \url{https://aclanthology.org/2021.emnlp-main.58}.

\bibitem[Rei et~al.(2020)Rei, Stewart, Farinha, and Lavie]{rei-etal-2020-unbabels}
Rei, R., Stewart, C., Farinha, A.~C., and Lavie, A.
\newblock Unbabel{'}s participation in the {WMT}20 metrics shared task.
\newblock In \emph{Proceedings of the Fifth Conference on Machine Translation}, pp.\  911--920, Online, November 2020. Association for Computational Linguistics.
\newblock URL \url{https://aclanthology.org/2020.wmt-1.101}.

\bibitem[Rei et~al.(2022)Rei, C.~de Souza, Alves, Zerva, Farinha, Glushkova, Lavie, Coheur, and Martins]{rei-etal-2022-comet}
Rei, R., C.~de Souza, J.~G., Alves, D., Zerva, C., Farinha, A.~C., Glushkova, T., Lavie, A., Coheur, L., and Martins, A. F.~T.
\newblock {COMET}-22: Unbabel-{IST} 2022 submission for the metrics shared task.
\newblock In \emph{Proceedings of the Seventh Conference on Machine Translation (WMT)}, pp.\  578--585, Abu Dhabi, United Arab Emirates (Hybrid), December 2022. Association for Computational Linguistics.
\newblock URL \url{https://aclanthology.org/2022.wmt-1.52}.

\bibitem[Robert \& Casella(1999)Robert and Casella]{robert1999monte}
Robert, C.~P. and Casella, G.
\newblock \emph{Monte Carlo statistical methods}, volume~2.
\newblock Springer, 1999.

\bibitem[Rush et~al.(2015)Rush, Chopra, and Weston]{rush-etal-2015-neural}
Rush, A.~M., Chopra, S., and Weston, J.
\newblock A neural attention model for abstractive sentence summarization.
\newblock In \emph{Proceedings of the 2015 Conference on Empirical Methods in Natural Language Processing}, pp.\  379--389, Lisbon, Portugal, September 2015. Association for Computational Linguistics.
\newblock \doi{10.18653/v1/D15-1044}.
\newblock URL \url{https://aclanthology.org/D15-1044}.

\bibitem[Sellam et~al.(2020)Sellam, Das, and Parikh]{sellam-etal-2020-bleurt}
Sellam, T., Das, D., and Parikh, A.
\newblock {BLEURT}: Learning robust metrics for text generation.
\newblock In \emph{Proceedings of the 58th Annual Meeting of the Association for Computational Linguistics}, pp.\  7881--7892, Online, July 2020. Association for Computational Linguistics.
\newblock \doi{10.18653/v1/2020.acl-main.704}.
\newblock URL \url{https://aclanthology.org/2020.acl-main.704}.

\bibitem[Stahlberg \& Byrne(2019)Stahlberg and Byrne]{stahlberg-byrne-2019-nmt}
Stahlberg, F. and Byrne, B.
\newblock On {NMT} search errors and model errors: Cat got your tongue?
\newblock In \emph{Proceedings of the 2019 Conference on Empirical Methods in Natural Language Processing and the 9th International Joint Conference on Natural Language Processing (EMNLP-IJCNLP)}, pp.\  3356--3362, Hong Kong, China, November 2019. Association for Computational Linguistics.
\newblock \doi{10.18653/v1/D19-1331}.
\newblock URL \url{https://aclanthology.org/D19-1331}.

\bibitem[Sutskever et~al.(2014)Sutskever, Vinyals, and Le]{Sutskever2014}
Sutskever, I., Vinyals, O., and Le, Q.~V.
\newblock Sequence to sequence learning with neural networks.
\newblock In \emph{Proceedings of the 27th International Conference on Neural Information Processing Systems - Volume 2}, NIPS'14, pp.\  3104–3112, Cambridge, MA, USA, 2014. MIT Press.

\bibitem[Suzgun et~al.(2023)Suzgun, Melas-Kyriazi, and Jurafsky]{suzgun-etal-2023-follow}
Suzgun, M., Melas-Kyriazi, L., and Jurafsky, D.
\newblock Follow the wisdom of the crowd: Effective text generation via minimum {B}ayes risk decoding.
\newblock In \emph{Findings of the Association for Computational Linguistics: ACL 2023}, pp.\  4265--4293, Toronto, Canada, July 2023. Association for Computational Linguistics.
\newblock \doi{10.18653/v1/2023.findings-acl.262}.
\newblock URL \url{https://aclanthology.org/2023.findings-acl.262}.

\bibitem[Welleck et~al.(2020)Welleck, Kulikov, Kim, Pang, and Cho]{welleck-etal-2020-consistency}
Welleck, S., Kulikov, I., Kim, J., Pang, R.~Y., and Cho, K.
\newblock Consistency of a recurrent language model with respect to incomplete decoding.
\newblock In \emph{Proceedings of the 2020 Conference on Empirical Methods in Natural Language Processing (EMNLP)}, pp.\  5553--5568, Online, November 2020. Association for Computational Linguistics.
\newblock \doi{10.18653/v1/2020.emnlp-main.448}.
\newblock URL \url{https://aclanthology.org/2020.emnlp-main.448}.

\bibitem[Wolf et~al.(2020)Wolf, Debut, Sanh, Chaumond, Delangue, Moi, Cistac, Rault, Louf, Funtowicz, Davison, Shleifer, von Platen, Ma, Jernite, Plu, Xu, Le~Scao, Gugger, Drame, Lhoest, and Rush]{wolf-etal-2020-transformers}
Wolf, T., Debut, L., Sanh, V., Chaumond, J., Delangue, C., Moi, A., Cistac, P., Rault, T., Louf, R., Funtowicz, M., Davison, J., Shleifer, S., von Platen, P., Ma, C., Jernite, Y., Plu, J., Xu, C., Le~Scao, T., Gugger, S., Drame, M., Lhoest, Q., and Rush, A.
\newblock Transformers: State-of-the-art natural language processing.
\newblock In \emph{Proceedings of the 2020 Conference on Empirical Methods in Natural Language Processing: System Demonstrations}, pp.\  38--45, Online, October 2020. Association for Computational Linguistics.
\newblock \doi{10.18653/v1/2020.emnlp-demos.6}.
\newblock URL \url{https://aclanthology.org/2020.emnlp-demos.6}.

\bibitem[Wu et~al.(2016)Wu, Schuster, Chen, Le, Norouzi, Macherey, Krikun, Cao, Gao, Macherey, Klingner, Shah, Johnson, Liu, Łukasz Kaiser, Gouws, Kato, Kudo, Kazawa, Stevens, Kurian, Patil, Wang, Young, Smith, Riesa, Rudnick, Vinyals, Corrado, Hughes, and Dean]{wu2016googles}
Wu, Y., Schuster, M., Chen, Z., Le, Q.~V., Norouzi, M., Macherey, W., Krikun, M., Cao, Y., Gao, Q., Macherey, K., Klingner, J., Shah, A., Johnson, M., Liu, X., Łukasz Kaiser, Gouws, S., Kato, Y., Kudo, T., Kazawa, H., Stevens, K., Kurian, G., Patil, N., Wang, W., Young, C., Smith, J., Riesa, J., Rudnick, A., Vinyals, O., Corrado, G., Hughes, M., and Dean, J.
\newblock Google's neural machine translation system: Bridging the gap between human and machine translation, 2016.

\bibitem[Xu et~al.(2024)Xu, Kim, Sharaf, and Awadalla]{xu2024paradigm}
Xu, H., Kim, Y.~J., Sharaf, A., and Awadalla, H.~H.
\newblock A paradigm shift in machine translation: Boosting translation performance of large language models.
\newblock In \emph{The Twelfth International Conference on Learning Representations}, 2024.
\newblock URL \url{https://openreview.net/forum?id=farT6XXntP}.

\bibitem[Yang et~al.(2023)Yang, Chen, Lin, and Byrne]{yang2023direct}
Yang, G., Chen, J., Lin, W., and Byrne, B.
\newblock Direct preference optimization for neural machine translation with minimum bayes risk decoding.
\newblock \emph{arXiv preprint arXiv:2311.08380}, 2023.

\bibitem[Zhang et~al.(2020)Zhang, Kishore, Wu, Weinberger, and Artzi]{bert-score}
Zhang, T., Kishore, V., Wu, F., Weinberger, K.~Q., and Artzi, Y.
\newblock {BERTS}core: Evaluating text generation with {BERT}.
\newblock In \emph{8th International Conference on Learning Representations, {ICLR} 2020, Addis Ababa, Ethiopia, April 26-30, 2020}. OpenReview.net, 2020.
\newblock URL \url{https://openreview.net/forum?id=SkeHuCVFDr}.

\end{thebibliography}

\clearpage

\appendix

\section{Proof of Theorem \ref{th:kld}}
\label{sec:proof}

From the convention $0 \log 0 = 0$ and the log sum inequality (LSI) (Theorem 2.7.1 in \citet{cover1999elements}), we have for any $p\in \Delta(\mathcal{Y}; \RefH)$:
\begin{align}
    D_{\mathrm{KL}}&(p || \Pm) \nonumber\\
            &\stackrel{0 \log 0 = 0}{=} \sum_{\vy \in \RefH} p(\vy) \log \frac{p(\vy)}{P(\vy)} \nonumber\\
            &\;\;\;\stackrel{LSI}{\geq} (\sum_{\vy \in \RefH} p(\vy)) \log \frac{\sum_{\vy \in \RefH} p(\vy)}{\sum_{\vy \in \RefH} P(\vy)} \nonumber\\
            &\;\;\;\;=-\log \sum_{\vy \in \RefH} P(\vy) \nonumber\\
            &\;\;\;\;=\log \alpha,
    \label{eq:lower_bound_on_kl}
\end{align}
where the third equality follows from the fact that $\sum_{\vy \in \RefH} p(\vy) = 1$, and the last equality follows from the definition of $\alpha = \frac{1}{\sum_{\vy \in \RefH} P(\vy)}$.

On the other hand, from \eqref{eq:mbp}, we get:
\begin{align*}
    D_{\mathrm{KL}}(\Pmb || \Pm) &= \sum_{\vy \in \RefH} \Pmb(\vy) \log \frac{p(\vy)}{P(\vy)} \\
    &= \sum_{\vy \in \RefH} \alpha P(\vy) \log \alpha \\
    &= (\alpha \log \alpha) \frac{1}{\alpha} = \log \alpha,
\end{align*}
which matches the lower bound in \eqref{eq:lower_bound_on_kl}.
Therefore, we have $D_{\mathrm{KL}}(p || \Pm) \geq D_{\mathrm{KL}}(\Pmb || \Pm)$ for any $p\in \Delta(\mathcal{Y}; \RefH)$.

\section{Proof of Lemma \ref{lem:pinsker}}
\label{sec:proof2}
Lemma \ref{lem:pinsker} is derived by Pinsker's inequality \cite{DBLP:books/cu/CsiszarK11}:
\begin{align*}
    |\mathop{\mathbb{E}}_{\vy \sim P}& [u(\vh, \vy)] - \mathop{\mathbb{E}}_{\vy \sim p} [u(\vh, \vy)]| \\
    &=|\sum_{\vy \in \mathcal{Y}} u(\vh, \vy) (P(\vy) - p(\vy))| \\
    &\leq u_{\mathrm{max}} \sum_{\vy \in \mathcal{Y}} |P(\vy) - p(\vy)| \\
    &\leq u_{\mathrm{max}} \sqrt{2 D_{\mathrm{KL}}(p || P)}.
\end{align*}

\section{Empirical Evaluation of the Divergence}
\label{sec:divergence}

\begin{figure}
    \centering
    \subfloat[IWSLT'17 Fr-En]{
    \includegraphics[width=0.95\columnwidth]{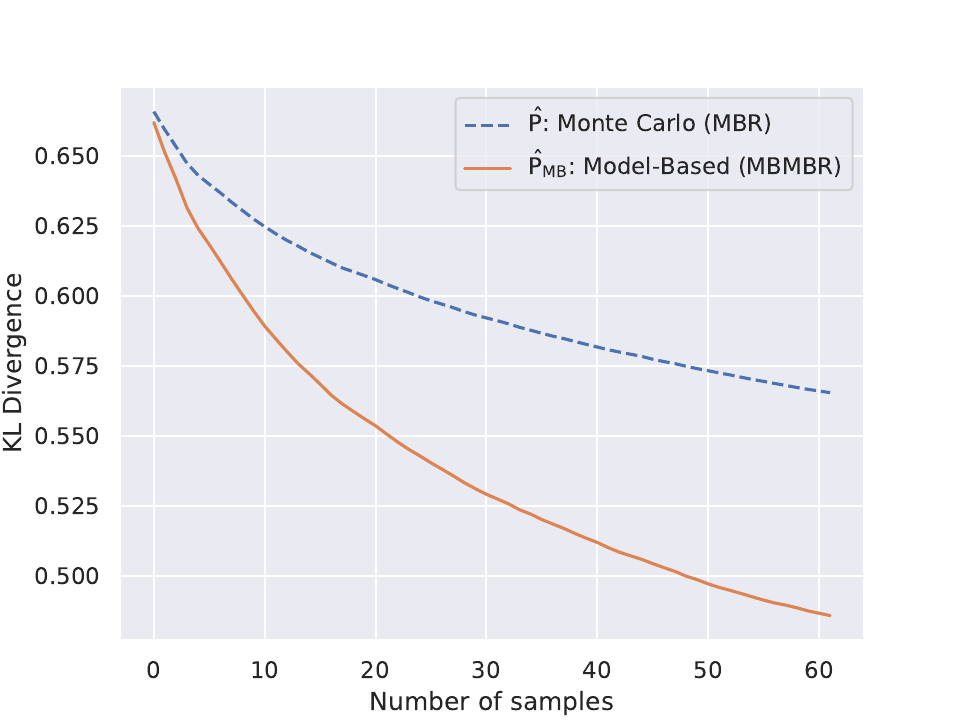}
    }\\
    \subfloat[XSum]{
    \includegraphics[width=0.95\columnwidth]{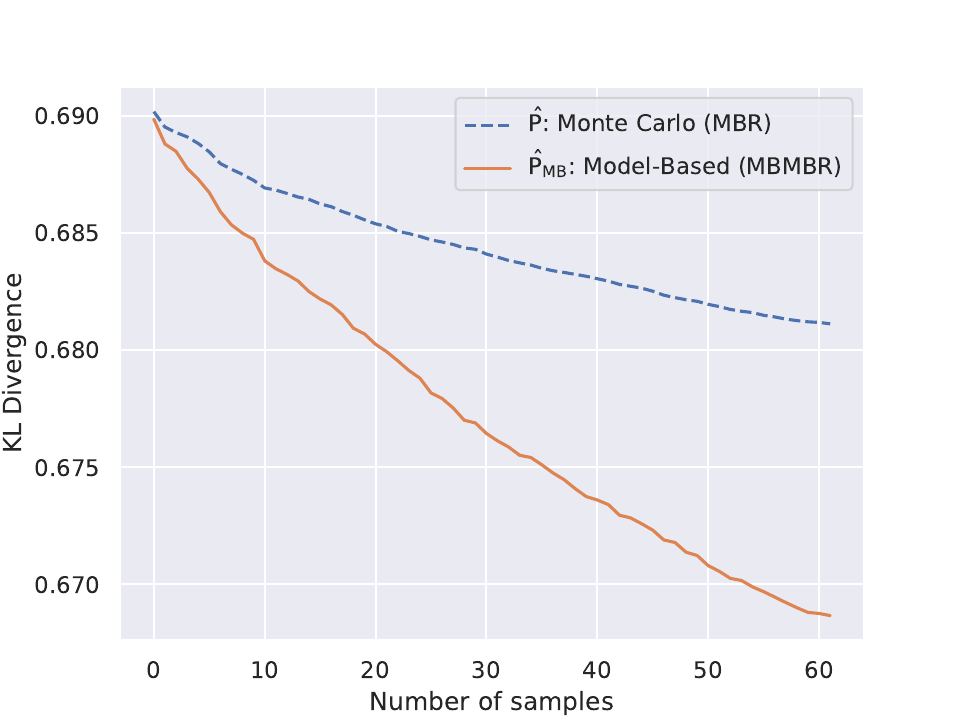}
    }\\
    \subfloat[MS COCO]{
    \includegraphics[width=0.95\columnwidth]{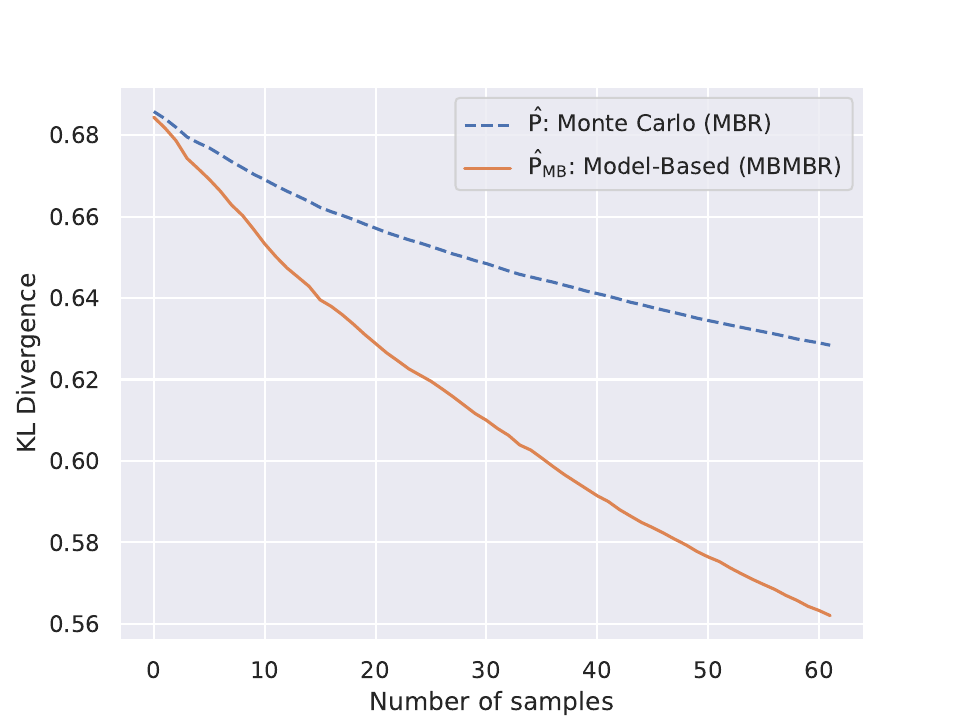}
    }
    \caption{Kullback-Leibler Divergence of the Monte Carlo estimate and the model-based estimate to the $\Pmodel$, averaged over the source sentences.}
    \label{fig:apx-kl}
\end{figure}

The result of the KL divergence on IWSLT'17 Fr-En, XSum, and MS COCO is shown in Figure \ref{fig:apx-kl}. See Section \ref{sec:analysis} for the experimental setup.
The model-based estimate significantly reduces the KL divergence compared to the Monte Carlo estimate in all domains.

\begin{figure*}[tb]
    \centering
    \subfloat[WMT'19 De-En]{
    \includegraphics[width=0.95\columnwidth]{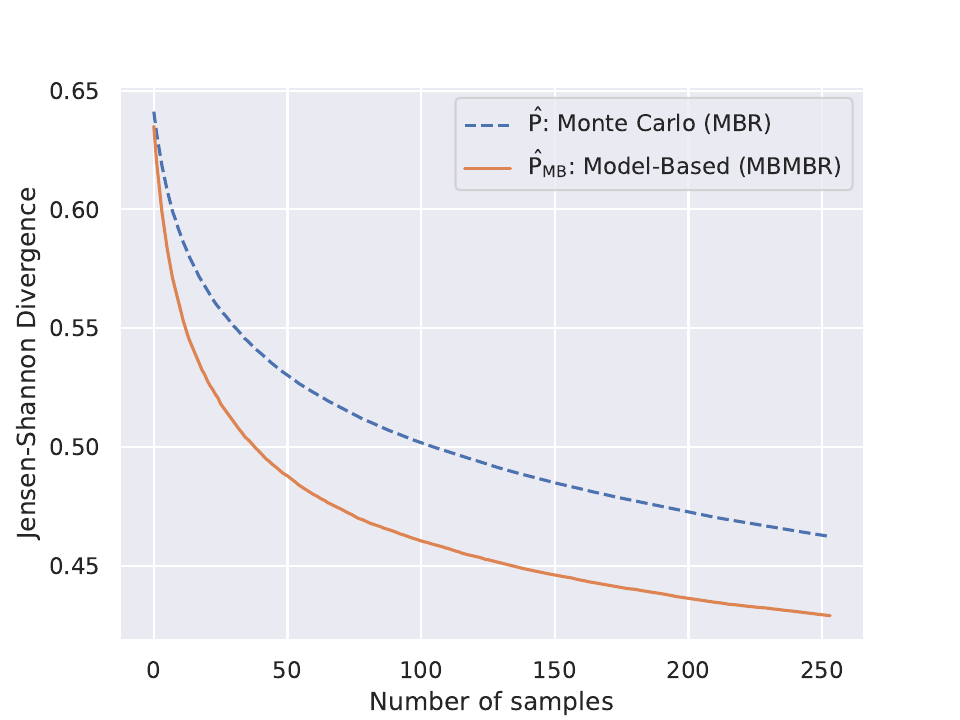}
    }
    \subfloat[IWSLT'17 Fr-En]{
    \includegraphics[width=0.95\columnwidth]{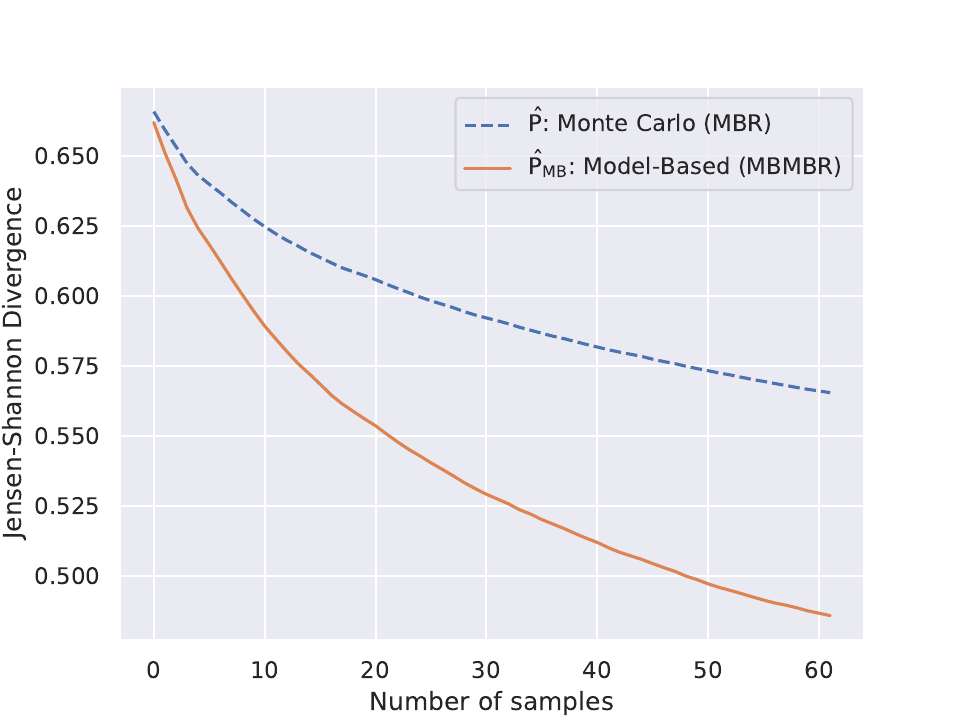}
    }\\
    \subfloat[XSum]{
    \includegraphics[width=0.95\columnwidth]{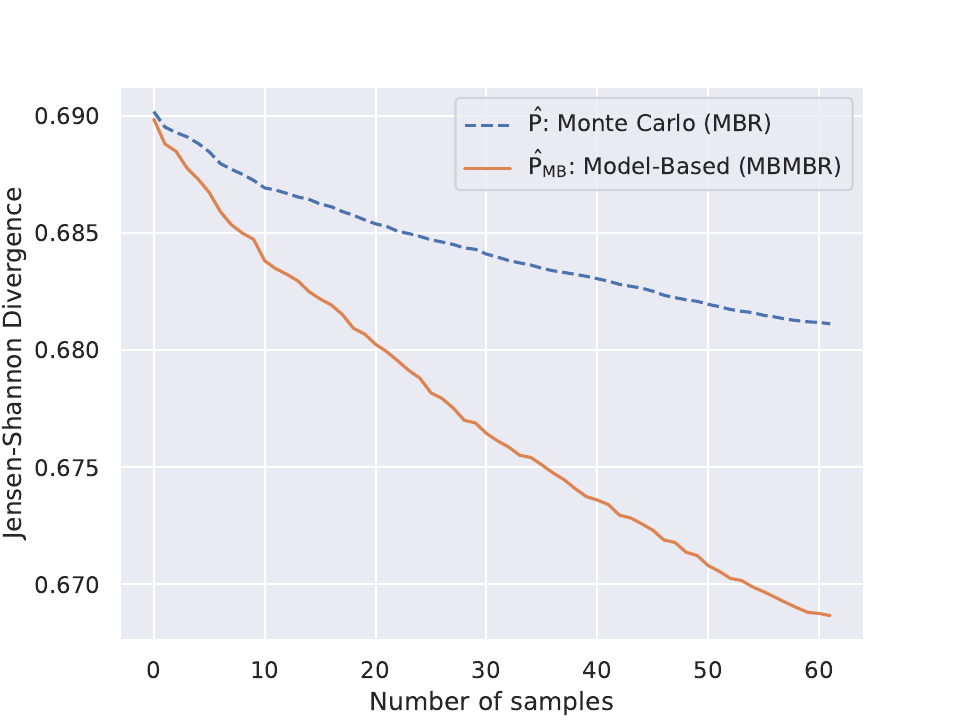}
    }
    \subfloat[MS COCO]{
    \includegraphics[width=0.95\columnwidth]{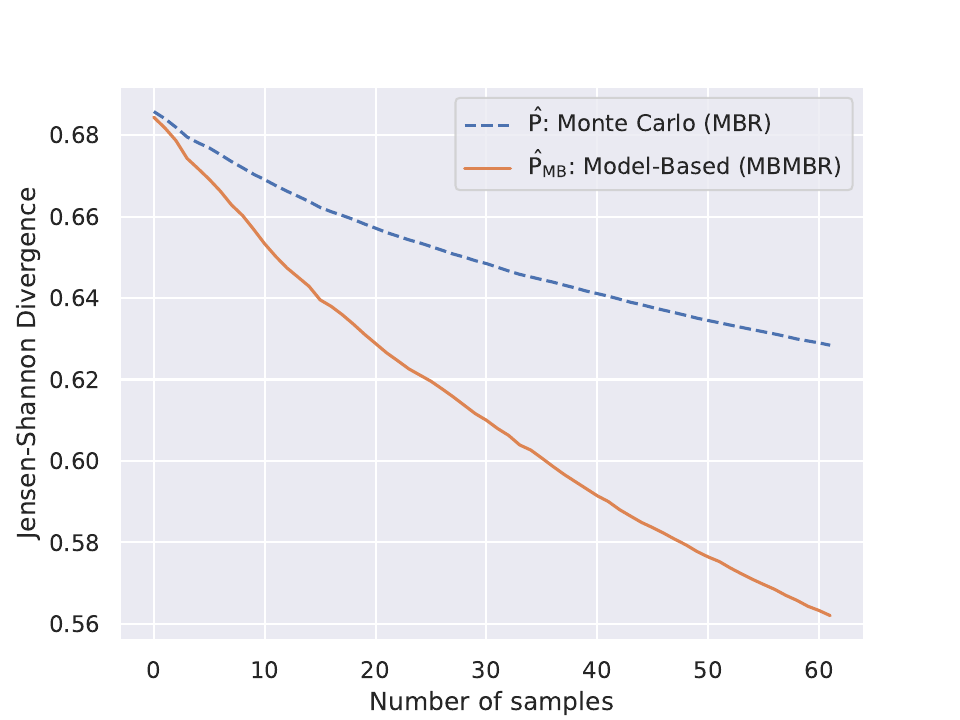}
    }
    \caption{Jensen-Shannon Divergence of the Monte Carlo estimate and the model-based estimate to the $\Pmodel$, averaged over the source sentences.}
    \label{fig:apx-jsd}
\end{figure*}

As an additional measure of the divergence, we compute the Jensen-Shannon Divergence (JSD). 
Note that the JSD of a distribution $p$ and $\Pm$ is computable without enumerating all the hypotheses in $\mathcal{Y}$ if the support of $p$ is restricted to $\RefH$. Let $M = \frac{1}{2} (p + P)$ be a mixture distribution of $p$ and $P$. Using the convention $0 \ln \frac{0}{q(x)} = 0$ \cite{cover1999elements}:
\begin{align*}
    J&SD(p || \Pm)\\
    =&\frac{1}{2} (D_{\mathrm{KL}}(p || M) + D_{\mathrm{KL}}(\Pm || M)) \\
    =& \frac{1}{2} (\sum_{\vy \in \mathcal{Y}} p(\vy) \ln(\frac{p(\vy)}{M(\vy)}) \\
     &+ \sum_{\vy \in \mathcal{Y}} \Pm(\vy) \ln(\frac{\Pm(\vy)}{M(\vy)})) \\
    =& \frac{1}{2} (\sum_{\vy \in \RefH} p(\vy) \ln(\frac{p(\vy)}{M(\vy)}) \\
     &+ \sum_{\vy \in \mathcal{Y} \setminus \RefH} p(\vy) \ln(\frac{p(\vy)}{M(\vy)}) \\
     &+ \sum_{\vy \in \RefH} \Pm(\vy) \ln(\frac{\Pm(\vy)}{M(\vy)})) \\
     &+ \sum_{\vy \in \mathcal{Y} \setminus \RefH} \Pm(\vy) \ln(\frac{\Pm(\vy)}{M(\vy)})) \\
    =& \frac{1}{2} (\sum_{\vy \in \RefH} p(\vy) \ln(\frac{p(\vy)}{M(\vy)}) \\
     &+ 0 \\
     &+ \sum_{\vy \in \RefH} \Pm(\vy) \ln(\frac{\Pm(\vy)}{M(\vy)})) \\
     &+ (1 - \sum_{\vy \in \RefH} \Pm(\vy)) \ln(2) ).
\end{align*}
Therefore, $JSD(p || \Pm)$ is computable without enumerating all the hypotheses in $\mathcal{Y}$.
Figure \ref{fig:apx-jsd} shows the JSD of the model-based and the Monte Carlo estimator. For all datasets, the JSD of the model-based distribution is significantly smaller than that of the empirical distribution. Compared to the Monte Carlo estimate, the model-based estimate requires less than half of the samples to achieve the same divergence.

\section{Evaluation using COMET and BLEURT}
\label{sec:comet}
The state-of-the-art evaluation function for neural machine translation is neural-based models \cite{freitag-etal-2022-high,fernandes-etal-2022-quality}. We evaluate using COMET-22 \cite{rei-etal-2020-unbabels,rei-etal-2022-comet} and BLEURT-20 \cite{sellam-etal-2020-bleurt,pu-etal-2021-learning} as a utility function for WMT'19 De-En and Ru-En using 64 samples with epsilon sampling ($\epsilon=0.02$). Other experimental settings are the same as in Section \ref{sec:mtm}.
The results are in Tables \ref{tab:comet22} and \ref{tab:bleurt}. MBMBR improves upon MBR using COMET-22 but decreases performance in De-En using BLEURT-20. 

\begin{table}
    \centering
    \begin{tabular}{ccc}
    \toprule
               & De-En & Ru-En \\\midrule
    \baseline	& 85.75 & 85.37\\
    \weighted & \textbf{85.84} & 85.44\\
    \lnw	& 85.82	&  \textbf{85.47} \\
    \bottomrule
    \end{tabular}
    \caption{Evaluation on WMT'19 using COMET-22 as the utility function and the evaluation metric.}
    \label{tab:comet22}
\end{table}

\begin{table}
    \centering
    \begin{tabular}{ccc}
    \toprule
     & De-En & Ru-En \\\midrule
    \baseline	& \textbf{28.31} & 31.14 \\
    \weighted & 26.92 & 29.42 \\
    \lnw	& 28.03 & \textbf{31.27}\\
    \bottomrule
    \end{tabular}
    \caption{Evaluation on WMT'19 using BLEURT-20 as the utility function and the evaluation metric.}
    \label{tab:bleurt}
\end{table}

\section{Evaluation with LLM-based Machine Translation Models}
\label{sec:tower}
Recent work shows that fine-tuning LLMs for tasks related to machine translation achieves state-of-the-art performance in machine translation.
We evaluate the performance of the MBMBR using two state-of-the-art LLM-based translation models, TowerInstruct-7B-v0.1 \cite{alves2024tower} and ALMA-7B-R \cite{xu2024paradigm}. We sample 64 outputs with epsilon sampling ($\epsilon=0.02$). The prompt for the models is described in Appendix \ref{sec:prompt}. Other experimental settings are the same as in Section \ref{sec:mt-llm}.
Table \ref{tab:llm-nmt} shows the COMET-22 and BLEU scores of the two models. MBMBR outperforms MBR using TowerInstruct-7B-v0.1 but slightly drops in performance using ALMA-7B-R.

\begin{table}
    \centering
    \begin{tabular}{ccc}
    \toprule
         & \multicolumn{2}{c}{TowerInstruct-7B-v0.1} \\
         \cmidrule(l){2-3}
         & COMET-22 & BLEU \\\midrule
        \baseline & 85.47 & 36.11 \\
        \weighted & \textbf{86.55} & 36.10 \\
        \lnw      & \textbf{86.55} & \textbf{36.35} \\\midrule\midrule
         & \multicolumn{2}{c}{ALMA-7B-R} \\
         \cmidrule(l){2-3}
         & COMET-22 & BLEU \\\midrule
        \baseline & \textbf{84.65} & \textbf{31.68} \\
        \weighted & 84.52 & 31.63 \\
        \lnw      & 84.39 & 31.46 \\
    \bottomrule
    \end{tabular}
    \caption{Evaluation of LLM-based translation models on IWSLT'17 Fr-En. BERTScore is used as a utility function.}
    \label{tab:llm-nmt}
\end{table}

\section{Evaluation of Sampling Algorithms for MBR Decoding}
\label{sec:eval-sampling}

We evaluate the performance of MBR with epsilon sampling ($\epsilon=0.02$), top-$k$ sampling ($k=10$), nucleus sampling ($p=0.9$), and ancestral sampling on text summarization and image captioning.
Other experimental settings are the same as in Section \ref{sec:experiments}.
The results are shown in Tables \ref{tab:xsum-sampling}, \ref{tab:samsum-sampling}, \ref{tab:mscoco-sampling}, and \ref{tab:nocaps-sampling}.
We observe that epsilon sampling outperforms the others on all four datasets, and its results are further improved with \weighted{} and \lnw{}.

\begin{table}[p]
    \centering
\adjustbox{max width=0.87\columnwidth}{
    \begin{tabular}{lrrrrr}
    \toprule
    \multicolumn{6}{c}{XSum} \\
    \midrule
     & \multicolumn{5}{c}{Epsilon Sampling ($\epsilon=0.02$)}  \\
     \cmidrule(l){2-6}
    $|\mathcal{H}|$ & 4 &  8 & 16 & 32 & 64 \\
    \midrule
    \baseline & 32.01 &  33.96 &   34.80 &  35.41 &   35.90  \\
    \weighted & 31.69 &  33.09 &   34.00 &   34.72 &   35.51  \\
    \lnw      & 32.65 &  34.01 &   35.00 &   35.63 &   36.18 \\
    \midrule
     & \multicolumn{5}{c}{Top-$k$ Sampling ($k=10$)} \\
     \cmidrule(l){2-6}
    $|\mathcal{H}|$ & 4 &  8 & 16 & 32 & 64 \\
    \midrule
    \baseline &  31.89 &  33.60 &   34.15 &   35.03 &   35.42 \\
    \weighted &  31.71 &  32.83 &   33.76 &   34.28 &   34.90 \\
    \lnw      &  32.28 &  33.76 &   34.38 &   35.21 &   35.83 \\
    \midrule
     & \multicolumn{5}{c}{Nucleus Sampling ($p=0.9$)} \\
     \cmidrule(l){2-6}
    $|\mathcal{H}|$ & 4 &  8 & 16 & 32 & 64  \\
    \midrule
    \baseline &  30.73 &  32.03 &   33.15 &   33.92 &   34.60 \\
    \weighted &  30.49 &  31.41 &   32.55 &   33.25 &   33.72 \\
    \lnw      &  30.69 &  31.79 &   32.96 &   33.96 &   34.68 \\
    \midrule
     & \multicolumn{5}{c}{Ancestral Sampling} \\
     \cmidrule(l){2-6}
    $|\mathcal{H}|$ & 4 &  8 & 16 & 32 & 64 \\
    \midrule
    \baseline &  27.48 &  28.27 &   29.41 &   30.60 &   31.82 \\
    \weighted &  24.72 &  25.88 &   26.97 &   28.63 &   30.16 \\
    \lnw      &  24.75 &  25.97 &   27.15 &   29.09 &   31.05 \\
    \bottomrule
    \end{tabular}
}
    \caption{ROUGE-L scores on XSum dataset. We evaluate using the first 1000 inputs of the dataset. BERTScore is used as the utility function. }
    \label{tab:xsum-sampling}
\end{table}

\begin{table}[p]
    \centering
\adjustbox{max width=0.87\columnwidth}{
    \begin{tabular}{lrrrrr}
    \toprule
    \multicolumn{6}{c}{SAMSum} \\
    \midrule
     & \multicolumn{5}{c}{Epsilon Sampling ($\epsilon=0.02$)}  \\
     \cmidrule(l){2-6}
    $|\mathcal{H}|$ & 4 &  8 & 16 & 32 & 64 \\
    \midrule
    \baseline & 30.55 &  31.16 &   31.98 &   32.51 &   33.03 \\
    \weighted & 31.68 &  32.44 &   33.14 &   33.86 &   34.50  \\
    \lnw      & 31.17 &  32.06 &   32.62 &   32.99 &   33.14 \\
    \midrule
     & \multicolumn{5}{c}{Top-$k$ Sampling ($k=10$)} \\
     \cmidrule(l){2-6}
    $|\mathcal{H}|$ & 4 &  8 & 16 & 32 & 64 \\
    \midrule
    \baseline &  28.04 &  28.21 &   28.17 &   28.33 &   31.81 \\
    \weighted &  28.32 &  28.91 &   29.84 &   30.76 &   31.34 \\
    \lnw      &  28.26 &  28.90 &   29.74 &   30.47 &   31.47 \\
    \midrule
     & \multicolumn{5}{c}{Nucleus Sampling ($p=0.9$)} \\
     \cmidrule(l){2-6}
    $|\mathcal{H}|$ & 4 &  8 & 16 & 32 & 64  \\
    \midrule
    \baseline &  27.76 &  27.84 &   28.13 &   28.24 &   30.95 \\
    \weighted &  27.87 &  28.21 &   28.32 &   28.68 &   29.17 \\
    \lnw      &  27.87 &  28.26 &   28.35 &   28.69 &   29.27 \\
    \midrule
     & \multicolumn{5}{c}{Ancestral Sampling} \\
     \cmidrule(l){2-6}
    $|\mathcal{H}|$ & 4 &  8 & 16 & 32 & 64 \\
    \midrule
    \baseline &  25.94 &  26.33 &   26.43 &   26.60 &   29.83 \\
    \weighted &  25.81 &  25.86 &   25.90 &   26.07 &   26.24 \\
    \lnw      &  25.81 &  25.86 &   25.90 &   26.07 &   26.26 \\
    \bottomrule
    \end{tabular}
}
    \caption{ROUGE-L scores on SAMSum dataset. BERTScore is used as the utility function.}
    \label{tab:samsum-sampling}
\end{table}

\begin{table}[p]
    \centering
\adjustbox{max width=0.87\columnwidth}{
    \begin{tabular}{lrrrrr}
    \toprule
    \multicolumn{6}{c}{MS COCO} \\
    \midrule
     & \multicolumn{5}{c}{Epsilon Sampling ($\epsilon=0.02$)}  \\
     \cmidrule(l){2-6}
    $|\mathcal{H}|$ & 4 &  8 & 16 & 32 & 64 \\
    \midrule
    \baseline & 27.12 & 28.46 & 30.42 & 31.51 & 33.24 \\
    \weighted & 28.75 & 30.56 & 33.12 & 34.25 & 34.96 \\
    \lnw      & 28.65 & 29.29 & 31.14 & 32.54 & 33.99 \\
    \midrule
     & \multicolumn{5}{c}{Top-$k$ Sampling ($k=10$)} \\
     \cmidrule(l){2-6}
    $|\mathcal{H}|$ & 4 &  8 & 16 & 32 & 64 \\
    \midrule
    \baseline & 26.13 & 28.08 & 30.07 & 31.89 & 32.40 \\
    \weighted & 21.36 & 23.80 & 26.22 & 28.81 & 30.93 \\
    \lnw      & 21.36 & 23.51 & 26.46 & 28.76 & 30.85 \\
    \midrule
     & \multicolumn{5}{c}{Nucleus Sampling ($p=0.9$)} \\
     \cmidrule(l){2-6}
    $|\mathcal{H}|$ & 4 &  8 & 16 & 32 & 64  \\
    \midrule
    \baseline & 26.40 & 29.12 & 31.12 & 31.88 & 32.98 \\
    \weighted & 21.68 & 24.36 & 26.84 & 29.08 & 31.00 \\
    \lnw      & 21.75 & 24.31 & 27.07 & 29.08 & 30.95 \\
    \midrule
     & \multicolumn{5}{c}{Ancestral Sampling} \\
     \cmidrule(l){2-6}
    $|\mathcal{H}|$ & 4 &  8 & 16 & 32 & 64 \\
    \midrule
    \baseline & 22.70 & 25.87 & 28.08 & 29.84 & 31.55 \\
    \weighted & 18.63 & 20.88 & 22.93 & 25.26 & 27.59 \\
    \lnw      & 18.59 & 20.48 & 22.65 & 25.90 & 29.52 \\
    \bottomrule
    \end{tabular}
}
    \caption{BLEU scores on MS COCO dataset. We evaluate using the first 1000 inputs of the dataset. BERTScore is used as the utility function.}
    \label{tab:mscoco-sampling}
\end{table}

\begin{table}[p]
    \centering
\adjustbox{max width=0.87\columnwidth}{
    \begin{tabular}{lrrrrr}
    \toprule
    \multicolumn{6}{c}{nocaps} \\
    \midrule
     & \multicolumn{5}{c}{Epsilon Sampling ($\epsilon=0.02$)}  \\
     \cmidrule(l){2-6}
    $|\mathcal{H}|$ & 4 &  8 & 16 & 32 & 64 \\
    \midrule
    \baseline & 25.43 & 27.67 & 28.73 & 30.93 & 31.42 \\
    \weighted & 23.31 & 24.73 & 25.87 & 26.91 & 27.27 \\
    \lnw      & 28.86 & 29.78 & 30.47 & 32.10 & 32.68 \\
    \midrule
     & \multicolumn{5}{c}{Top-$k$ Sampling ($k=10$)} \\
     \cmidrule(l){2-6}
    $|\mathcal{H}|$ & 4 &  8 & 16 & 32 & 64 \\
    \midrule
    \baseline & 23.69 & 25.91 & 27.41 & 29.33 & 30.67 \\
    \weighted & 21.39 & 23.01 & 24.05 & 25.18 & 25.66 \\
    \lnw      & 26.64 & 28.63 & 29.15 & 31.53 & 31.58 \\
    \midrule
     & \multicolumn{5}{c}{Nucleus Sampling ($p=0.9$)} \\
     \cmidrule(l){2-6}
    $|\mathcal{H}|$ & 4 &  8 & 16 & 32 & 64  \\
    \midrule
    \baseline & 21.58 & 25.76 & 27.20 & 28.55 & 29.24 \\
    \weighted & 20.89 & 22.91 & 25.29 & 26.70 & 27.14 \\
    \lnw      & 25.48 & 28.72 & 29.72 & 30.56 & 31.43 \\
    \midrule
     & \multicolumn{5}{c}{Ancestral Sampling} \\
     \cmidrule(l){2-6}
    $|\mathcal{H}|$ & 4 &  8 & 16 & 32 & 64 \\
    \midrule
    \baseline & 18.48 & 21.54 & 23.93 & 24.89 & 25.61 \\
    \weighted & 17.42 & 19.80 & 22.35 & 23.40 & 23.66 \\
    \lnw      & 21.40 & 25.10 & 27.57 & 28.47 & 29.36 \\
    \bottomrule
    \end{tabular}
}
    \caption{BLEU scores on nocaps dataset. We evaluate using the first 1000 inputs of the dataset. BERTScore is used as the utility function.}
    \label{tab:nocaps-sampling}
\end{table}

\section{Prompts for Large Language Models}
\label{sec:prompt}
For IWSLT'17 in Section \ref{sec:mt-llm}, We use the following prompt as a guide to trigger its translation capability for BLOOMZ and mT0:
\begin{quote}
    Translate the following sentence from French to English. \\
    Q: [[QUESTION]] \\
    A:
\end{quote}
We use the following prompt for TowerInstruct-7B-v0.1 and ALMA-7B-R in Appendix \ref{sec:tower}:
\begin{quote}
    Translate the following text from French into English.\\
    French: [[QUESTION]]\\
    English:
\end{quote}

For CNN/DM, we use the following prompt:

\begin{quote}
Given a BBC article, write a short summary of the article in one sentence. \\
Article: [[QUESTION]] \\
Q: Summarize the above article briefly in one sentence. \\
A:
\end{quote}

For E2E NLG, we use the few-shot learning prompt provided by \citet{suzgun-etal-2023-follow}.\footnote{\url{https://github.com/suzgunmirac/crowd-sampling/blob/3f7fc674925d5e43691b3e23def0bd3b6b0ff799/prompts/e2e_nlg_clean_fs.txt}}

\section{Additional Figures}
\label{sec:figures}
\begin{figure*}
    \centering
    \subfloat[WMT'19 De-En (Epsilon)]{
    \includegraphics[width=0.32\textwidth]{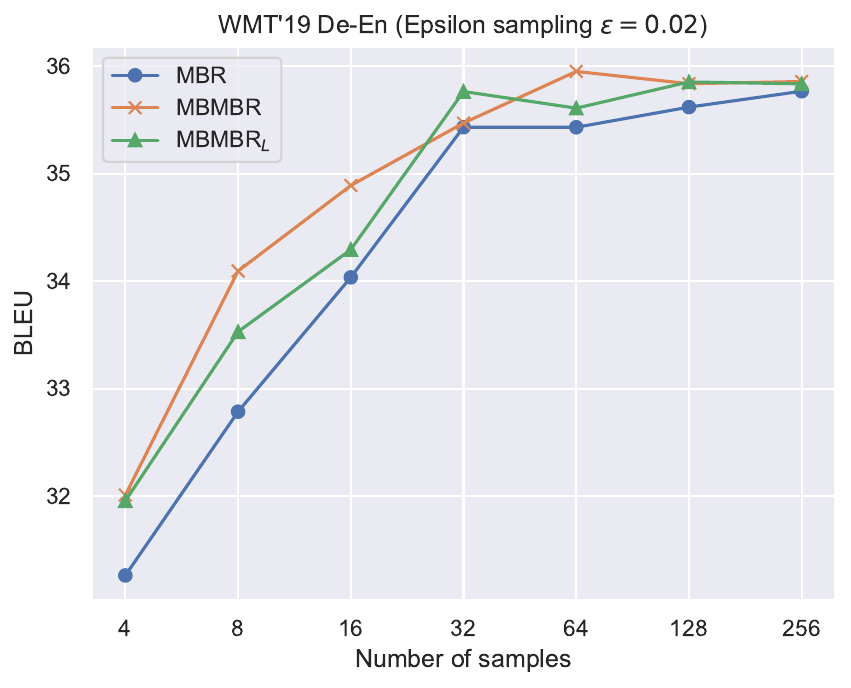}
    }%
    \subfloat[XWMT'19 Ru-En (Epsilon)]{
    \includegraphics[width=0.32\textwidth]{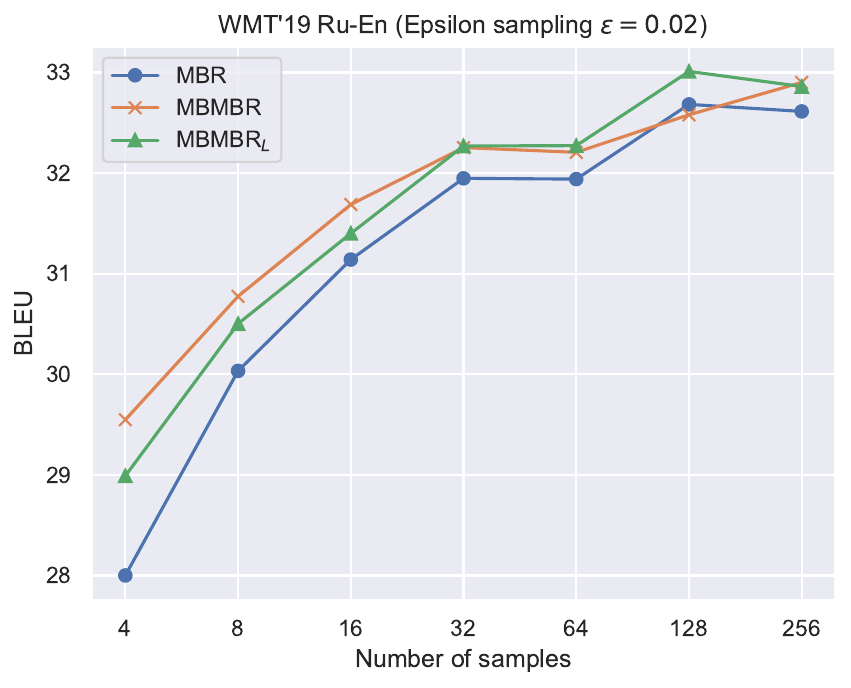}
    }\\
    \subfloat[WMT'19 De-En (Top-$k$)]{
    \includegraphics[width=0.32\textwidth]{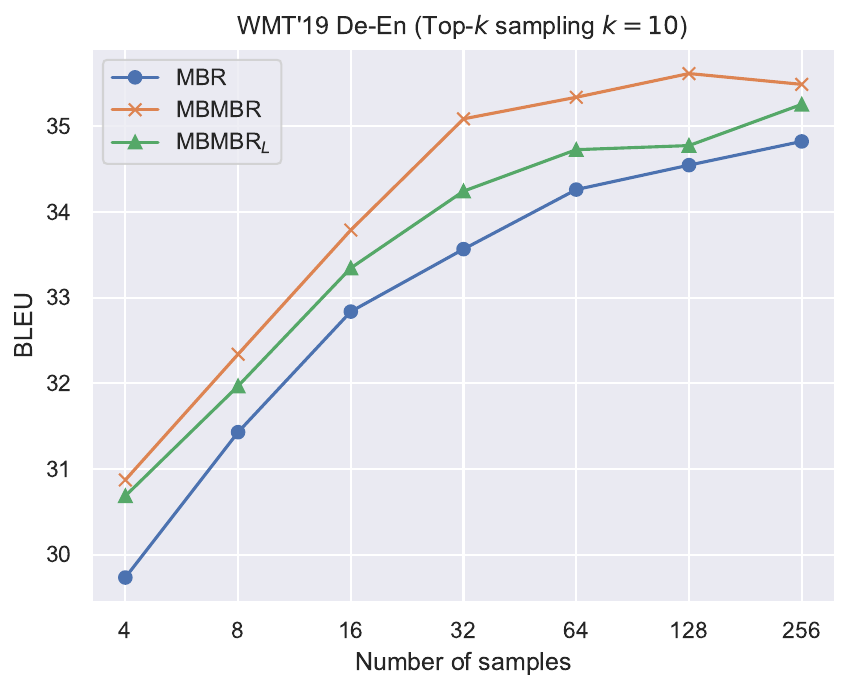}
    }%
    \subfloat[XWMT'19 Ru-En (Top-$k$)]{
    \includegraphics[width=0.32\textwidth]{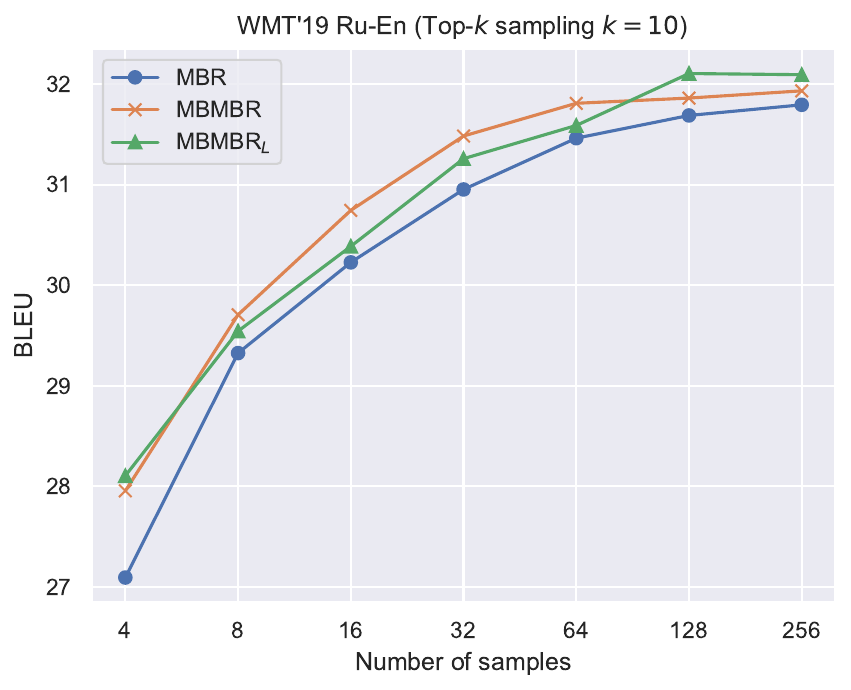}
    }\\
    \subfloat[WMT'19 De-En (Nucleus)]{
    \includegraphics[width=0.32\textwidth]{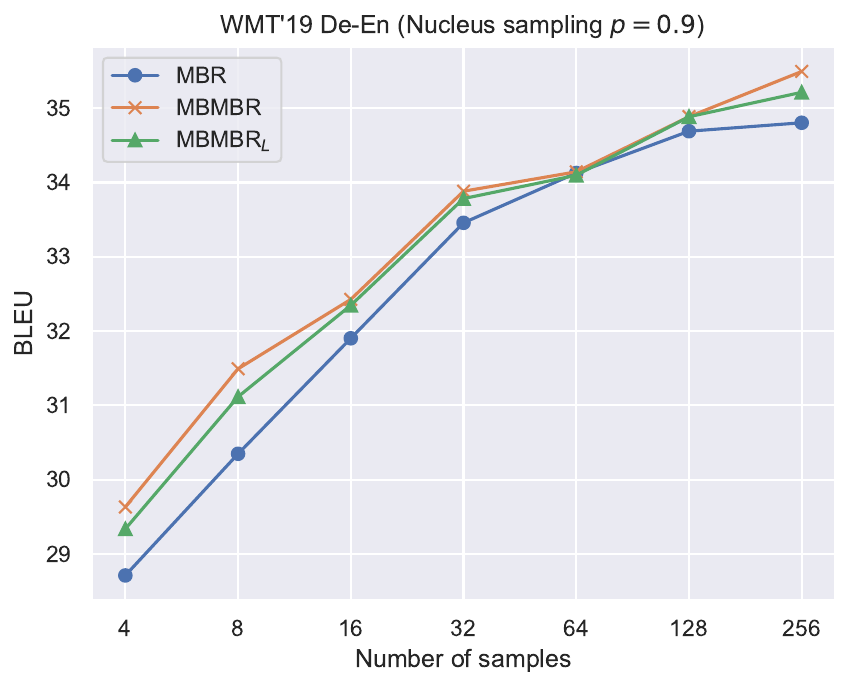}
    }%
    \subfloat[XWMT'19 Ru-En (Nucleus)]{
    \includegraphics[width=0.32\textwidth]{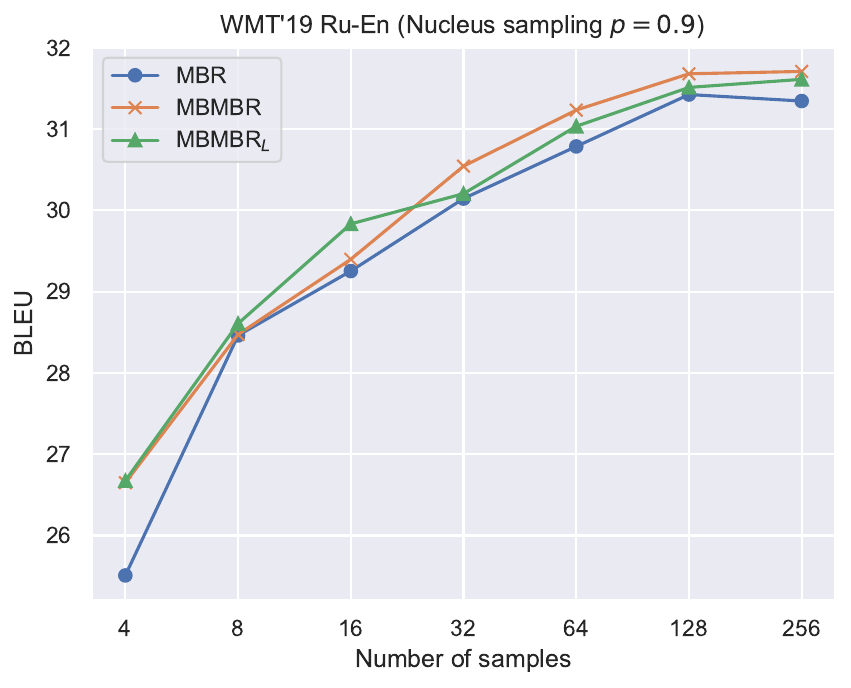}
    }\\
    \subfloat[WMT'19 De-En (Ancestral)]{
    \includegraphics[width=0.32\textwidth]{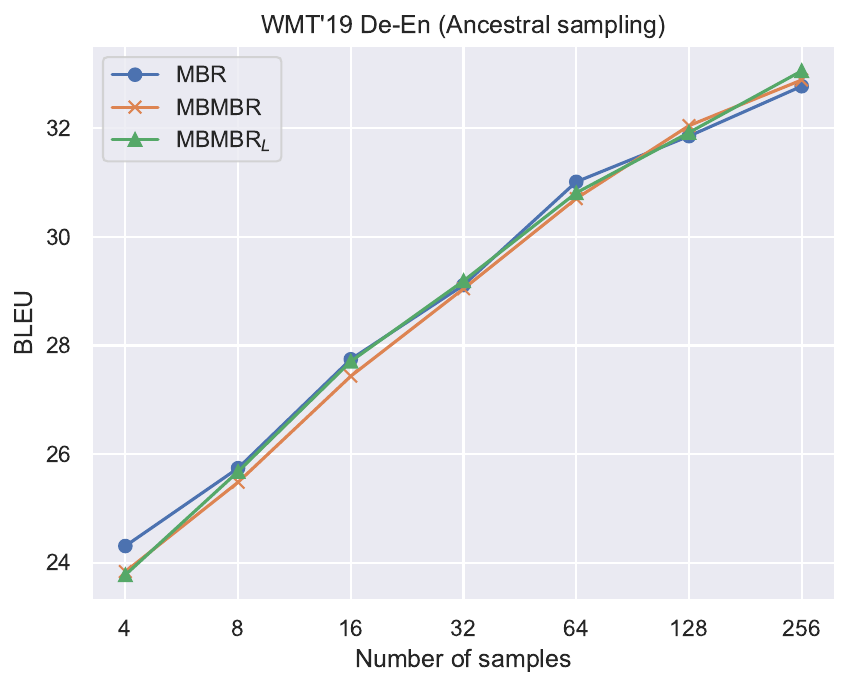}
    }%
    \subfloat[XWMT'19 Ru-En (Ancestral)]{
    \includegraphics[width=0.32\textwidth]{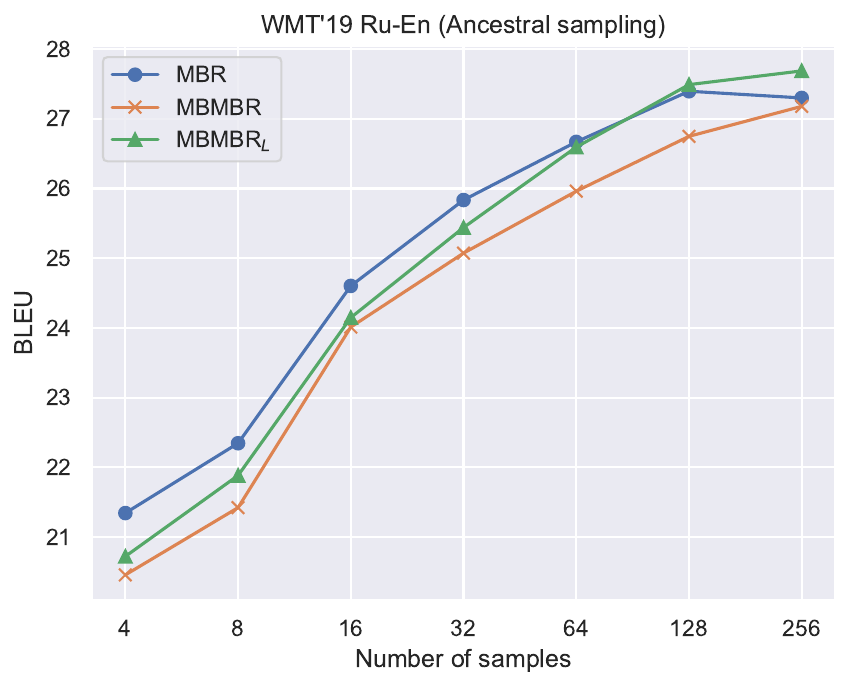}
    }
    \caption{BLEU score as a function of the number of samples on WMT'19 De-En and Ru-En (Section~\ref{sec:mtm}).}
    \label{fig:wmt19x-en}
\end{figure*}

\begin{figure*}
   \begin{minipage}{0.62\textwidth}
    \centering
    \subfloat[WMT'19 En-De (Epsilon)]{
    \includegraphics[width=0.5\textwidth]{figures/scale/wmt19.de-en-epsilon.pdf}
    }%
    \subfloat[XWMT'19 En-Ru (Epsilon)]{
    \includegraphics[width=0.5\textwidth]{figures/scale/wmt19.ru-en-epsilon.pdf}
    }
    \caption{BLEU score as a function of the number of samples on WMT'19 En-De and En-Ru (Section~\ref{sec:mtm}).}
    \label{fig:wmt19en-x}
   \end{minipage}\hfill
   \begin{minipage}{0.31\textwidth}
    \includegraphics[width=1\textwidth]{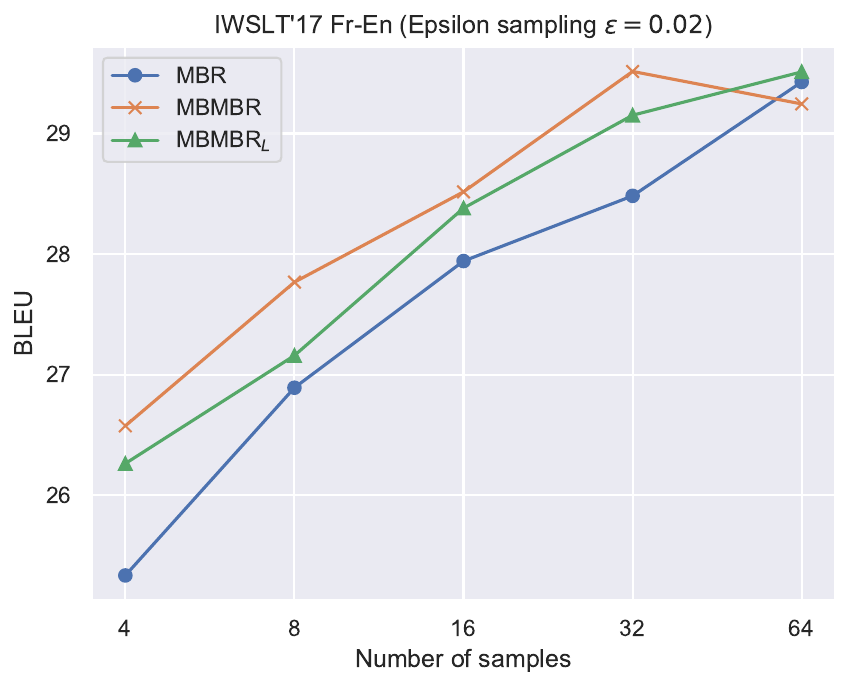}
    \caption{BLEU score as a function of the number of samples on IWSLT'17 Fr-En (Section~\ref{sec:mt-llm}).}
    \label{fig:iwslt}
   \end{minipage}
\end{figure*}

\begin{figure*}
   \begin{minipage}{0.62\textwidth}
    \centering
    \subfloat[XSum (Epsilon)]{
    \includegraphics[width=0.5\textwidth]{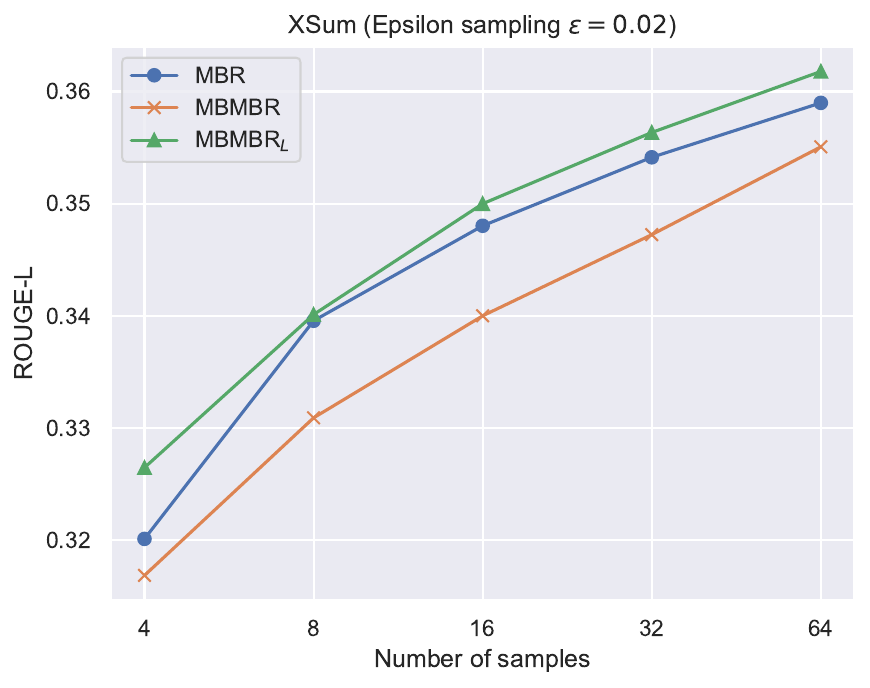}
    }%
    \subfloat[SAMSum (Epsilon)]{
    \includegraphics[width=0.5\textwidth]{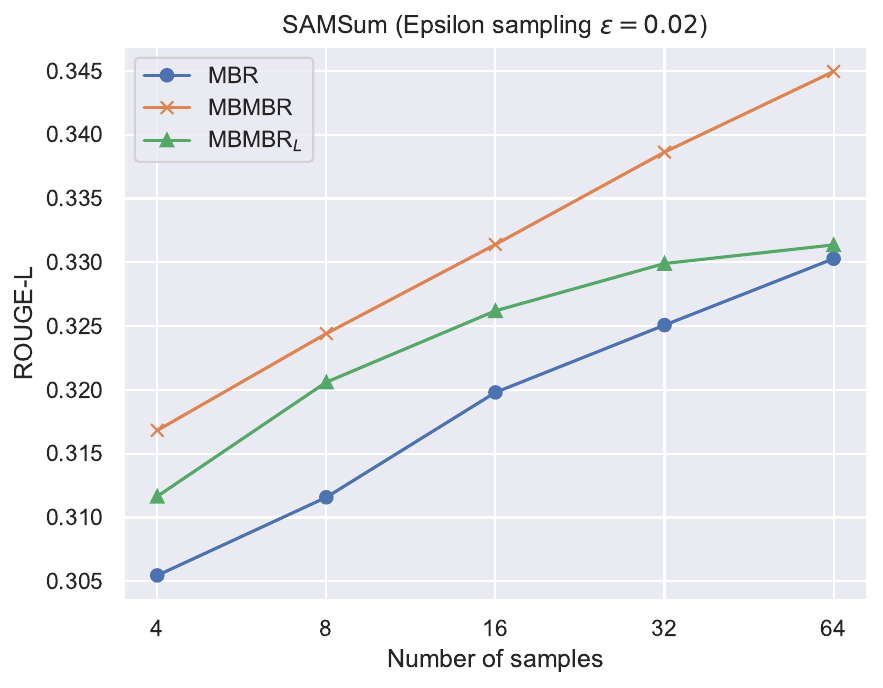}
    }
    \caption{ROUGE-L score as a function of the number of samples on XSum and SAMSum (Section~\ref{sec:sumsum}).}
    \label{fig:sum}
   \end{minipage}\hfill
   \begin{minipage}{0.31\textwidth}
    \centering
    \includegraphics[width=\textwidth]{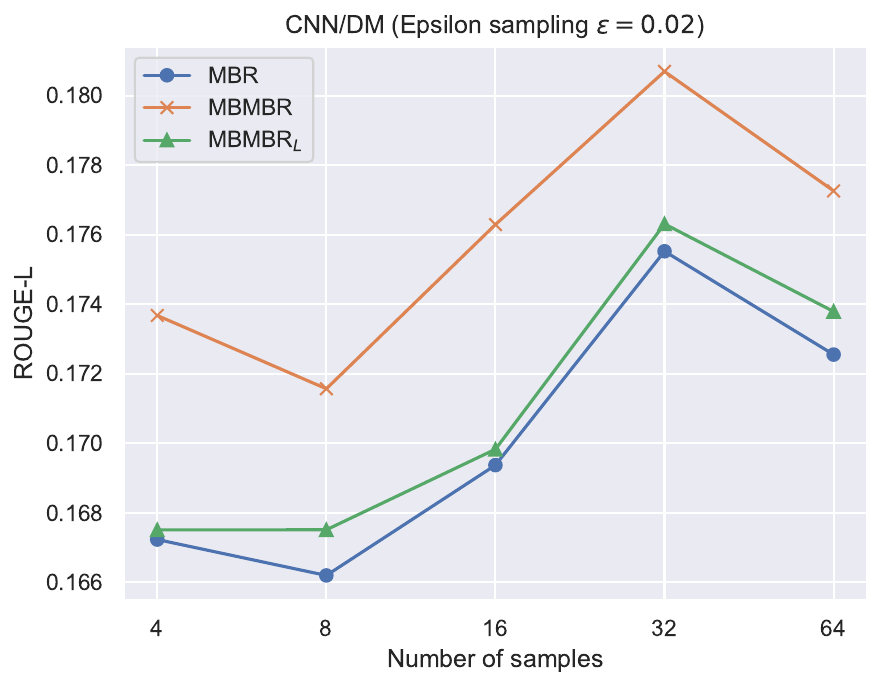}
    \caption{ROUGE-L score as a function of the number of samples on CNN/DM (Section~\ref{sec:sum-llm}).}
    \label{fig:cnndm}
   \end{minipage}
\end{figure*}

\begin{figure*}
\begin{minipage}{0.62\textwidth}
    \centering
    \subfloat[MSCOCO (Epsilon)]{
    \includegraphics[width=0.5\textwidth]{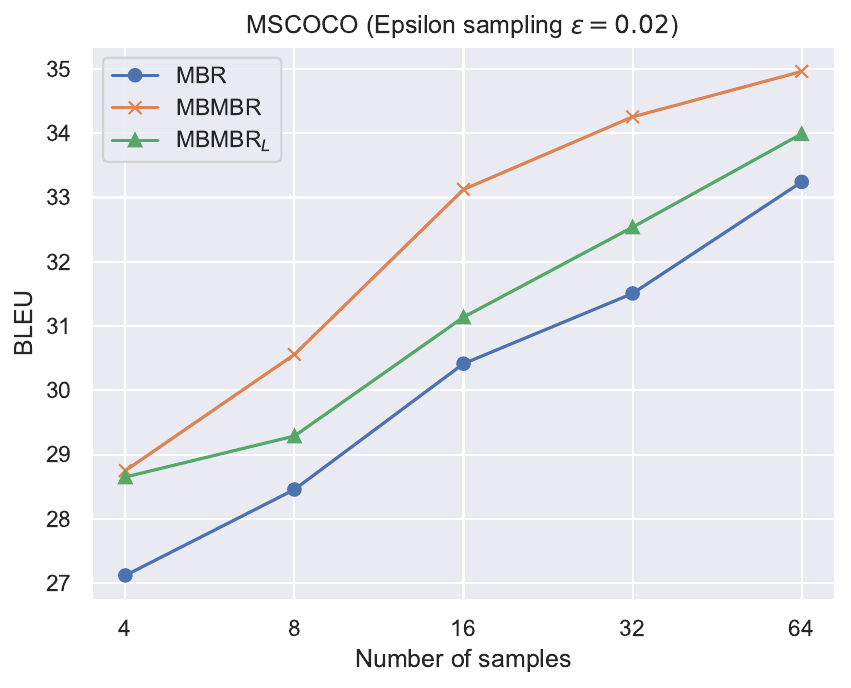}
    }%
    \subfloat[nocaps (Epsilon)]{
    \includegraphics[width=0.5\textwidth]{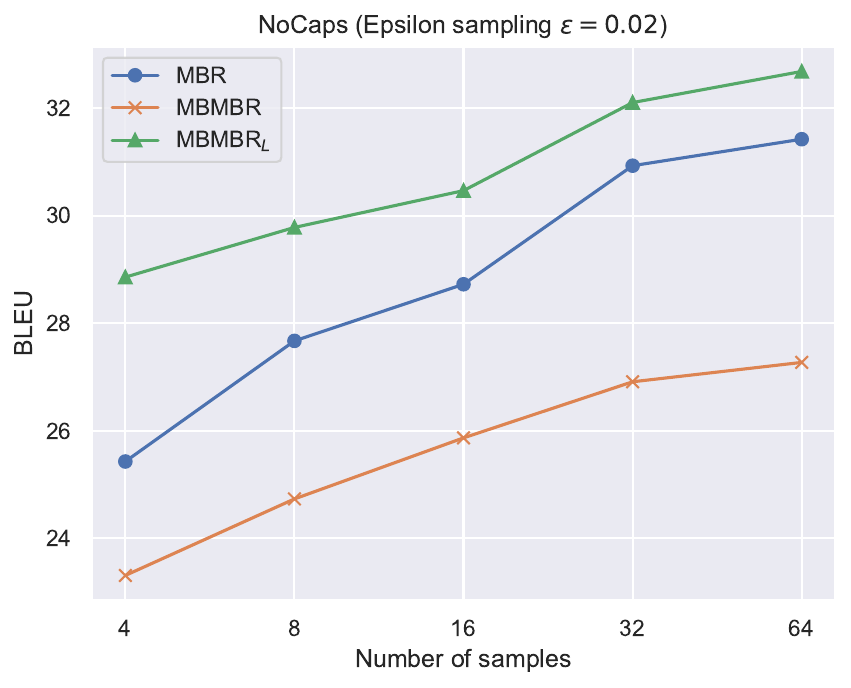}
    }
    \caption{BLEU score as a function of the number of samples on MSCOCO and nocaps (Section~\ref{sec:captioning}).}
    \label{fig:caps}
\end{minipage}\hfill
\begin{minipage}{0.31\textwidth}
    \centering
    \includegraphics[width=1\textwidth]{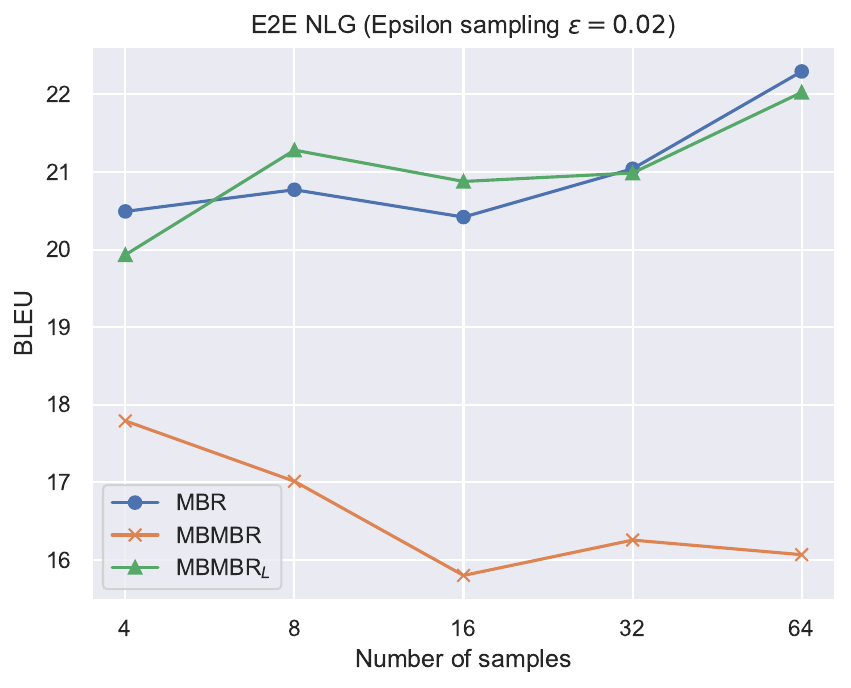}
    \caption{BLEU score as a function of the number of samples on E2E NLG (Section~\ref{sec:e2e}).}
    \label{fig:e2e}
\end{minipage}
\end{figure*}

Figures \ref{fig:wmt19x-en}, \ref{fig:wmt19en-x}, \ref{fig:iwslt}, \ref{fig:sum}, \ref{fig:cnndm}, \ref{fig:caps}, \ref{fig:e2e} show the performance of the methods as a function of the number of samples.

\section{Pretrained Models used in the Experiments}

We list the pretrained models we used in the experiments in Table \ref{tab:models}.

\begin{table*}
    \centering
    \adjustbox{max width=\textwidth}{
    \begin{tabular}{cl}
    \toprule
    Task & Model \\
    \midrule
        WMT'19 (Section \ref{sec:mtm}) & \citet{ng-etal-2019-facebook} \url{https://github.com/facebookresearch/fairseq/blob/main/examples/wmt19/README.md} \\
        IWSLT 2017 (Section \ref{sec:mt-llm}) & \citet{muennighoff-etal-2023-crosslingual} \url{https://huggingface.co/bigscience/bloomz-7b1-mt} \\
        XSum (Section \ref{sec:sumsum}) & \citet{lewis-etal-2020-bart} \url{https://huggingface.co/facebook/bart-large-xsum} \\
        SAMSum (Section \ref{sec:sumsum}) & \url{https://huggingface.co/philschmid/bart-large-cnn-samsum} \\
        CNN/DM (Section \ref{sec:sum-llm}) & \citet{jiang2023mistral} \url{https://huggingface.co/mistralai/Mistral-7B-Instruct-v0.1} \\
        MS COCO (Section \ref{sec:captioning}) & \citet{pmlr-v202-li23q} \url{https://huggingface.co/Salesforce/blip2-flan-t5-xl-coco} \\
        nocaps (Section \ref{sec:captioning}) & \citet{pmlr-v202-li23q} \url{https://huggingface.co/Salesforce/blip2-flan-t5-xl} \\
        E2E NLG (Section \ref{sec:e2e}) & \citet{jiang2023mistral} \url{https://huggingface.co/mistralai/Mistral-7B-Instruct-v0.1} \\
        \multirow{2}{*}{IWSLT 2017 (Appendix \ref{sec:tower})} & \citet{alves2024tower} \url{https://huggingface.co/Unbabel/TowerInstruct-7B-v0.1}\\
        & \citet{xu2024paradigm} \url{https://huggingface.co/haoranxu/ALMA-7B-R}\\
    \bottomrule
    \end{tabular}
    }
    \caption{List of pretrained models we used in the experiments.}
    \label{tab:models}
\end{table*}

\end{document}